\documentclass[lettersize,journal]{IEEEtran}
\usepackage{amsmath,amsfonts}
\usepackage{algorithm}
\usepackage{algpseudocode}
\usepackage{array}
\usepackage[caption=false,font=normalsize,labelfont=sf,textfont=sf]{subfig}
\usepackage{textcomp}
\usepackage{stfloats}
\usepackage{url}
\usepackage{verbatim}
\usepackage{graphicx}
\usepackage{booktabs}
\usepackage{cite}
\usepackage{hyperref}
\usepackage{adjustbox}
\usepackage{threeparttable}
\usepackage[dvipsnames]{xcolor}
\usepackage{multirow}
\definecolor{mypink}{rgb}{0.858, 0.188, 0.478}
\definecolor{myred}{RGB}{219, 0, 0}
\definecolor{mygray}{gray}{0.6}

\begin{document}

\title{DECODE: Domain-aware Continual Domain Expansion for Motion Prediction}

\author{Boqi Li\textsuperscript{*}, Haojie Zhu\textsuperscript{*}, Henry X. Liu\textsuperscript{*\dag},~\IEEEmembership{IEEE Fellow}
\thanks{*: Boqi Li, Haojie Zhu, and Henry X. Liu are with the Department of Civil and Environmental Engineering, University of Michigan, Ann Arbor, MI, USA. (e-mails: \tt\small [boqili, zhuhj, henryliu]@umich.edu)}
\thanks{\dag: corresponding author}
\thanks{© 2026 IEEE.  Personal use of this material is permitted.  Permission from IEEE must be obtained for all other uses, in any current or future media, including reprinting/republishing this material for advertising or promotional purposes, creating new collective works, for resale or redistribution to servers or lists, or reuse of any copyrighted component of this work in other works.}
\thanks{DOI: 10.1109/TPAMI.2026.3683469}
}

\markboth{Journal of \LaTeX\ Class Files,~Vol.~14, No.~8, August~2021}%
{Shell \MakeLowercase{\textit{et al.}}: A Sample Article Using IEEEtran.cls for IEEE Journals}


\maketitle

\begin{abstract}
Motion prediction is essential for autonomous vehicles to navigate complex environments and anticipate the behavior of other traffic participants. As new driving scenarios emerge, models must be continually updated without retraining from scratch. We propose DECODE, a continual learning framework that starts from a pre-trained generalized model and incrementally expands specialized models for distinct domains. Unlike existing approaches that pursue a single unified model, DECODE explicitly balances specialization and generalization through dynamic model selection. It employs a hypernetwork for parameter generation, which reduces storage costs, and utilizes a normalizing flow for real-time domain inference via likelihood estimation. Outputs from specialized and generalized models are fused using Bayesian uncertainty estimation. This integration ensures optimal performance in familiar conditions while maintaining robustness in novel scenarios. Extensive experiments show DECODE achieves a low forgetting rate of 0.044 and an average minADE of 0.584 m, outperforming prior methods and generalizing well across diverse driving domains. Furthermore, we demonstrate that DECODE can be extended beyond motion prediction to general continual learning tasks such as image classification, showcasing its broad applicability.
\end{abstract}


\section{Introduction}
\IEEEPARstart{M}{otion} prediction has attracted significant attention in recent years, finding applications across a wide array of fields. It plays a critical role in the development of autonomous vehicles, enabling them to understand complex scenarios and anticipate the future behaviors of other traffic participants \cite{huang2023differentiable, cheng2024pluto}. This capability allows for more informed decision-making, which is crucial not only for enhancing roadway safety but also for improving traffic efficiency and reducing congestion. Beyond road transport, motion prediction is instrumental in areas such as robotics for navigation and interaction \cite{foka2010probabilistic, fridovich2020confidence}, and in simulation environments where predictive models support the development and testing of various autonomous systems \cite{9561666, yan2023learning}. The task of motion prediction is inherently challenging due to the complexity of interactions among target agents and the multi-modal nature of their possible future movements.

\begin{figure}[!t]
\centering
\includegraphics[width=3.2in]{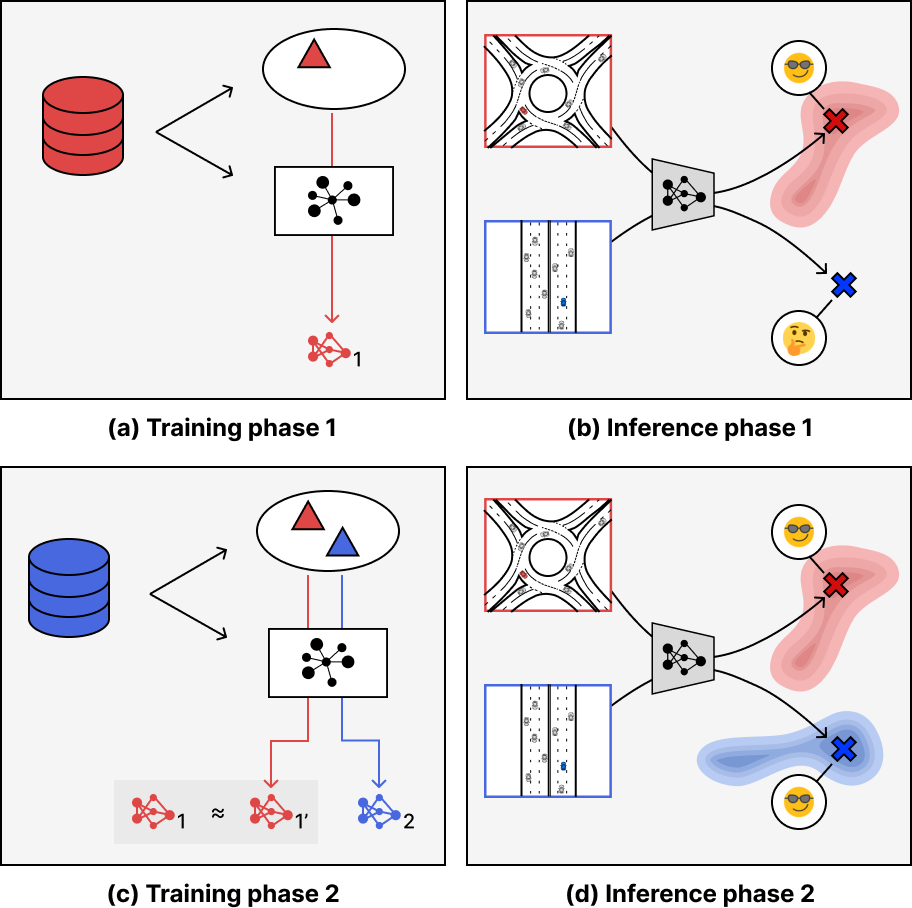}
\caption{Overview of DECODE framework: (a) The initial dataset updates the hypernetwork and creates a domain query to generate parameters for specialized model 1. (b) During inference, scenarios similar to the first domain achieve high likelihood scores from the normalizing flow, prompting the use of specialized model 1; those outside the domain have low likelihoods, leading to the use of the generalized model. (c) A new dataset is available, creating another domain query while ensuring consistent parameters for previous queries. (d) During subsequent inference, previously unfamiliar scenarios that now gain higher likelihood scores will utilize specialized model 2.}
\label{fig_1}
\end{figure}

Recent advancements in state-of-the-art algorithms have focused on integrating complex multi-agent interactions and map data \cite{salzmann2020trajectron++, zhao2021tnt, gu2021densetnt, shi2024mtr++}, yielding successful performances on several prominent open datasets \cite{Chang_2019_CVPR, zhan2019interaction, ettinger2021large}. However, these models require extensive datasets covering a broad spectrum of interactions involving multiple agents and road topology. Recent studies \cite{feng2024unitraj} have shown that without further training, models that excel on one dataset may perform poorly on others, posing a challenge for autonomous vehicles' global deployment and operation in diverse, unforeseen scenarios. Traditionally, motion prediction development has involved training models on entire datasets—a method still prevalent in current state-of-the-art efforts. However, the nature of autonomous driving development, characterized by the continuous acquisition of new and varied driving scenarios, necessitates frequent updates through model retraining. Retraining on all data combined becomes progressively more resource-intensive, while retraining solely on new data can lead to the loss of previously acquired knowledge \cite{wu2022continual}. This situation underscores the critical need for ongoing innovation in motion prediction methodologies and the development of effective continual learning strategies to enhance vehicle safety and reliability across varied environments. Therefore, in this paper we propose a novel continual learning framework for motion prediction called Domain-awarE COntinual Domain Expansion (DECODE). 

The DECODE framework starts from a pre-trained generalized model and incrementally develops a set of specialized models, each tailored to a specific domain where further adaption from the generalized model is beneficial. Unlike most existing continual learning approaches \cite{bao2023lifelong, feng2023continual, wu2022continual, ma2021continual} for motion predictions which attempt to develop a unified model capable of generalizing across diverse scenarios, we argue that a model that accumulates experience and adapts to specific domains, differentiated by characteristics such as various daily routines, road topologies, weather conditions, and local social norms would be more beneficial. A generalized model, trained to encompass a wide array of scenarios, might dilute the nuances of localized settings due to blended dataset biases. On the other hand, the necessity for generalization remains, as there will always be some common behaviors learned by a generalized model that we can rely on even when the users encounter unfamiliar driving conditions. The DECODE framework is unique in the way that it maintains a set of models that balance specialization with generalization, dynamically adjusting based on real-time requirements. As new datasets from different domains become available, we expand the overarching model by integrating additional specialist models. This strategy, while intuitive, introduces key research questions that must be addressed: how to expand the model without causing knowledge forgetting or excessive memory consumption, how to efficiently determine the appropriate specialist model to deploy in real-time, and how to manage scenarios that no existing specialist model can adequately handle.

To address the questions outlined, we first introduce an innovative model architecture that utilizes a hypernetwork \cite{ha2017hypernetworks, Oswald2020Continual} to dynamically generate parameters for each specialized model, significantly reducing the storage burden common in model expansion approaches. Instead of accumulating new parameters with each learning iteration, our method requires only a single set of parameters for the hypernetwork, as is shown in Fig. \ref{fig_1}. Second, the DECODE framework incorporates a normalizing flow \cite{rezende2015variational} mechanism for likelihood estimation during real-world inference. This allows the selection of the specialized model that best matches the current scenario based on the highest likelihood estimate, optimizing familiarity and relevance. Finally, when the available specialized models do not suffice for a given scenario, DECODE maintains a baseline performance through a generalized model pre-trained on diverse data. We integrate outputs from this generalized model and the most relevant specialized model using deep Bayesian uncertainty estimation techniques \cite{charpentier2020posterior, charpentier2022natural}, ensuring robust predictions across various conditions.

Extensive experiments confirm the effectiveness and superior performance of the DECODE framework. By leveraging a hypernetwork for continual learning, we achieved a notably low forgetting rate of 0.044, surpassing several other active learning strategies, and an average minADE of just 0.584 meters. Our normalizing flow-based model identification method accurately distinguishes among different domains, providing robust evidence of its ability to align model selection with specific operational scenarios, achieving an AUROC of 0.988. We further validated our framework using both naturalistic trajectory data and synthetically generated data, demonstrating its capacity to consistently produce high-quality motion predictions across a diverse range of domains, including safety-critical scenarios, with an accuracy of 0.986. Although the framework is developed with motion prediction in mind, its core design is not task-specific, and we demonstrate its potential applicability to general continual learning problems through additional experiments.  

Our contributions are as follows: (1) We propose a novel motion prediction framework that utilizes a hypernetwork to incrementally train a set of specialized models. This innovative architecture not only effectively reduces the issue of catastrophic forgetting but also manages to maintain an ensemble of models without increasing the storage burden. (2) We introduce a normalizing flow model that eliminates the need for explicit domain labels, demonstrating through experimental results its ability to accurately determine the appropriate specialized model during inference. (3) We incorporate the principles of deep Bayesian uncertainty estimation to merge predictions from both generalized and specialized models. This approach ensures robust outputs that are always bounded by the reliability of the generalized model, while simultaneously leveraging the specialized model's adeptness at handling familiar scenarios. 

\section{Related Work} 
\subsection{Motion Prediction}
Neural network-based learning approaches have significantly outperformed traditional model-based methods such as constant velocity models and Kalman filters in motion prediction \cite{lefevre2014survey}. These neural networks typically employ an encoder-decoder structure. During the encoding phase, techniques such as recurrent neural networks and 1D convolutions take advantage of the sequential nature of data to encode past trajectories. Recent research has further advanced motion prediction by focusing on two critical scene contexts: multi-agent interactions and local map features. Initial pioneering efforts, including Social LSTM \cite{Alahi_2016_CVPR} and Social GANs \cite{Gupta_2018_CVPR}, explored the impact of multi-agent interactions on future behaviors. Meanwhile, local maps have typically been rasterized into 2D images \cite{chai2019multipath, tang2019multiple, salzmann2020trajectron++}, allowing CNNs to extract detailed scene representations from them.

Subsequent advances in Graph Neural Networks (GNNs) and transformers have refined the capture of scene context information. LaneGCN \cite{liang2020learning} effectively discretizes the local map into lane nodes along lane centerlines, creating a connected lane graph that utilizes graph convolutions to extract relational information among agents and lanes—an approach echoed by several later models \cite{zeng2021lanercnn, gilles2022gohome}. Additionally, VectorNet \cite{gao2020vectornet}, widely adopted in recent research \cite{zhao2021tnt, gu2021densetnt, 9812107}, encodes both agents and map elements as polylines into a global graph. A graph neural network is then employed to encode this graph, effectively integrating the spatial relationships within the scene. Originally utilized in natural language processing, transformers have recently gained significant traction in motion prediction \cite{liu2021multimodal, girgis2021latent, huang2022multi, shi2024mtr++}. Like graph neural networks, they can flexibly extract relational information among agents and map elements by leveraging the multi-head attention mechanism \cite{NIPS2017_3f5ee243}.

Given the possibility of multiple future outcomes, many studies have incorporated multi-modal predictions to approximate the distribution of potential future motions of a target agent. These approaches typically generate a discrete set of future predictions, each with a corresponding confidence score to capture outcome variability. Early methods \cite{tang2019multiple , liang2020learning} used multiple regression heads to produce unique modes. Generative models like GANs \cite{Gupta_2018_CVPR, Sadeghian_2019_CVPR} and CVAEs\cite{ salzmann2020trajectron++, Chen_2022_CVPR} are also widely employed. More recent methods \cite{Phan-Minh_2020_CVPR, Narayanan_2021_CVPR, pmlr-v87-casas18a} often involve an intermediate classification phase to identify a distinct future mode, followed by regression to refine the motion. For example, TNT \cite{zhao2021tnt} and denseTNT \cite{gu2021densetnt} use agent's target endpoints to represent future modes, GOHOME \cite{gilles2022gohome} predicts future motions from a heatmap, and MTR++ \cite{shi2024mtr++} employ a learnable intention query with k-means clustered intention points. Our DECODE framework is compatible with any motion prediction model that utilizes an encoder-decoder structure and is flexible regarding input format, allowing for the integration of state-of-the-art scene context encoding methods. Furthermore, our framework adeptly handles the mult-modality of predictions by acknowledging the potential for multiple viable future trajectories and is particularly effective with methods that incorporate an intermediate classification phase.

\subsection{Continual Learning}
Continual Learning is predicated on the assumption that data arrives sequentially. The core challenge in this field is maintaining a balance between stability and plasticity—essentially, how to integrate new knowledge while preserving existing information. This equilibrium is vital as learning from new domains may result in catastrophic forgetting, a phenomenon where previously acquired knowledge is lost.

Continual learning can be broadly categorized into four approaches. Parameter isolation methods, such as progressive neural networks \cite{ rusu2016progressive}, expand untouched model parameters with the arrival of new domain data, completely protecting previous knowledge at the cost of increased model complexity and parameter storage. Regulation-based methods mitigate forgetting by adding a regularization term to the loss function \cite{aljundi2018memory}; an example is Elastic Weight Consolidation (EWC) \cite{ kirkpatrick2017overcoming}, which employs Fisher information to protect important parameters related to old knowledge during training.  Knowledge distillation-based methods, such as Learning without Forgetting (LwF) \cite{ li2017learning}, leverage outputs from old models to distill knowledge into the new model, preserving prior learning while integrating new information. Lastly, rehearsal-based methods include experience replay, which utilizes a selective collection from older datasets, and generative replay, which uses synthetically generated samples from previously learned models.

Recently, continual learning with pre-trained models has gained increasing attention. Traditional continual learning methods train models from scratch, which is inefficient and overlooks the prior knowledge embedded in pre-trained models. Building upon transfer learning, parameter-efficient tuning (PET) methods—such as prompt/prefix tuning \cite{lester2021power,li2021prefix}, adapter tuning \cite{houlsby2019parameter, rebuffi2017learning}, and LoRA \cite{hu2022lora} —adapt large pre-trained models with only a small number of learnable parameters. These techniques have been extended to continual learning to mitigate catastrophic forgetting while maintaining efficiency. Representative examples include L2P \cite{wang2022learning} and DualPrompt \cite{wang2022dualprompt}, which maintain prompt pools and use query-based selection, and CODA-Prompt \cite{smith2023coda}, which decomposes prompts and employs attention-based assembly for end-to-end optimization. Side-tuning \cite{zhang2020side} extends adapter-based methods but needs task identity during inference, and \cite{gao2023unified} offers a unified PET framework for continual learning.

Continual learning, well-established in computer vision fields like image classification, is newly emerging in motion prediction with a few recent initiatives. Most methods employ rehearsal-based methods: Ma et al. \cite {ma2021continual} enhanced generative replay with context from an experience replay buffer; Yang et al. \cite{ yang2022continual} combined experience replay with an external memory module for pedestrian trajectory prediction. Feng et al. \cite{ feng2023continual} integrated experience and generative replay with an uncertainty-aware module for selective sample generation. Wu et al. \cite{ wu2022continual}  developed a scene-level generative replay, using crowd interactions to generate trajectory scenes, while Knoedler et al. \cite{knoedler2022improving} merged experience replay with EWC. Differing from typical CVAE-based approaches, Bao et al. \cite{ bao2023lifelong}  proposed a CGAN-based generative replay for synthetic sample generation across multiple domains.

These existing approaches present several drawbacks. Experience replay struggles with selecting representative samples and necessitates external storage, which can quickly become unmanageable \cite{ parisi2019continual}. Generative replay's effectiveness heavily relies on the quality of synthetic data, posing significant challenges for state-of-the-art motion prediction models that require handling complex inputs like varying agent numbers and intricate road geometry. Moreover, training one AI model using synthetically generated data from another can degrade performance, a risk highlighted by Shumailov et al. in a recent Nature study \cite{shumailov2024ai}. To address these issues, our DECODE framework employs hypernetworks \cite{Oswald2020Continual}, which innovatively combine parameter isolation with the regularization of loss function. Similar to parameter-efficient tuning methods, our approach begins with a pre-trained generalized model and incrementally expands a set of specialized models based on it, enabling efficient adaptation while preserving prior knowledge. This approach efficiently preserves the integrity of previous models’ parameters while integrating new knowledge, offering a scalable and effective solution to the challenges of continual learning in motion prediction.

\section{DECODE for motion prediction}
\begin{figure*}[!t]
      \centering
      \includegraphics[trim=0 50 0 50, clip, width=7.0in]{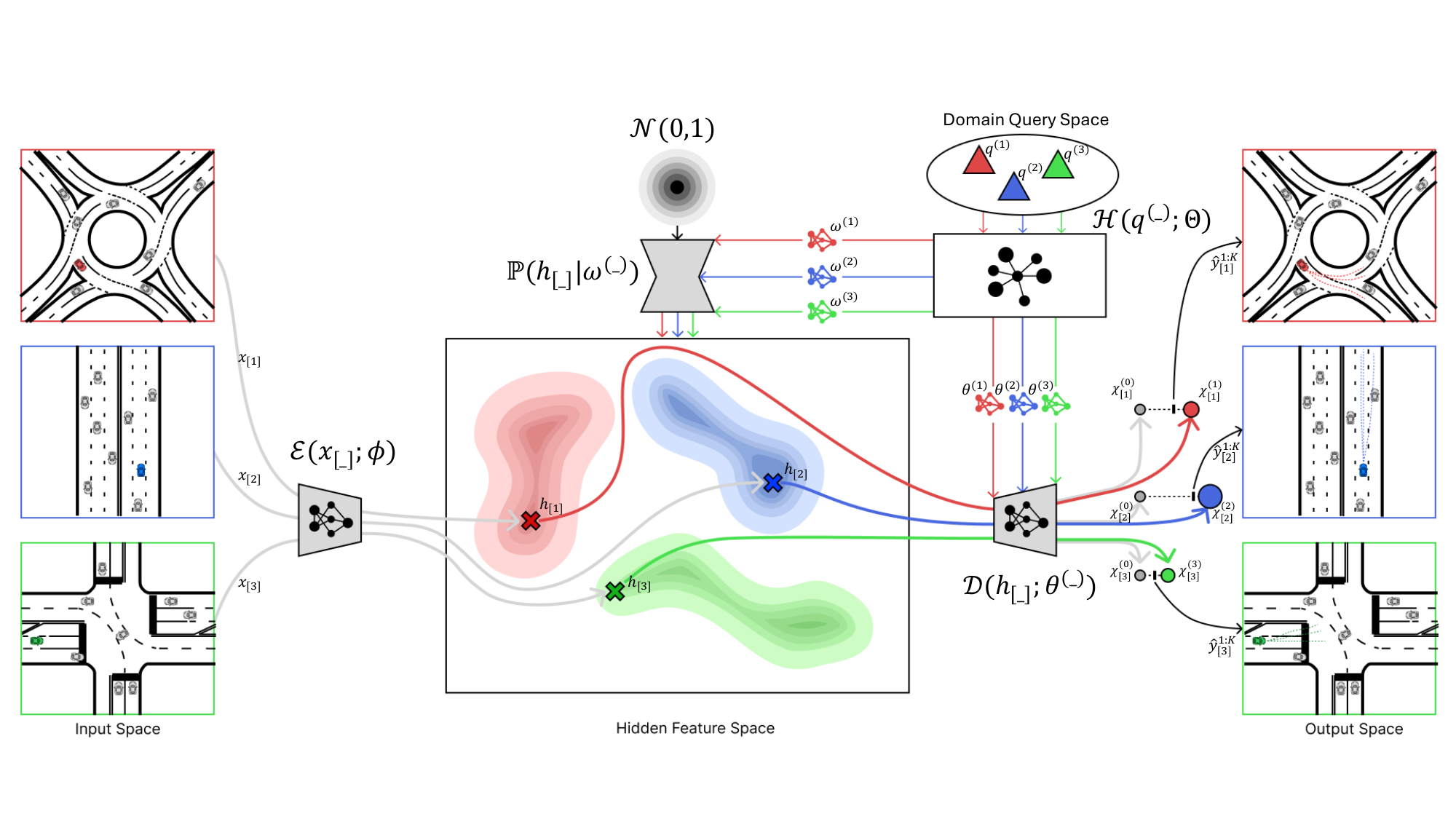}
      \caption{Overall Framework of DECODE: A set of domain queries serve as inputs to a hypernetwork, which dynamically generates parameters for both the normalizing flow and the decoder. The normalizing flow models, parameterized differently for each domain, generate distributions for domain-specific features. These distributions help identify the most suitable specialized model for each scene, based on the hidden representations produced by the encoder from the input space. The selected specialized decoder then generates future motion predictions for that scene. Outputs from the generalized model are integrated to ensure that the final prediction remains reliable and within performance bounds.} 
      \label{fig:framework}
\end{figure*}

The overall framework of DECODE is illustrated in Fig. \ref{fig:framework}. Our proposed learning strategy is agnostic to the specific motion prediction algorithm, requiring only that it adheres to an encoder-decoder structure. We begin with a pre-trained generalized model, which has been trained on a large dataset encompassing diverse scenarios. Continual learning is initiated when new domain data are introduced, and a decrease in the pre-trained model’s performance suggests the need for retraining. The subsequent paragraphs detail the implementation of our methodology: Section \ref{sec:hyper} introduces the hypernetwork used for incremental domain expansion. This is followed by Section \ref{sec:nf}, which describes the normalizing flow that provides our framework with the capability to discern which specialized model best fits the current scenario during inference. Finally, Section \ref{sec:evid} discusses how outputs from the generalized and specialized models are combined using deep Bayesian uncertainty estimation to ensure a robust performance boundary.
\subsection{Hypernetworks for Continual Domain Expansion} \label{sec:hyper}
In the task of motion prediction, given past observations of a target agent denoted as $ s_P = [s_{-T_P+1}, s_{-T_P+2}, \ldots, s_0]$, the objective is to predict its future trajectories, represented as \( y = [s_1, s_2, \ldots, s_{T_F}] \), where \( T_P \) and \( T_F \) respectively specify the number of historical and future timesteps. The agent's future behavior largely depends on its interactions with neighboring entities and road map information, collectively denoted as context \( c \). The total input is denoted as $x=(s_P, c)$ Despite the diversity of techniques in prior works, most motion prediction algorithms can be conceptualized as having an encoder-decoder structure. The encoder $\mathcal{E}$ processes the input to generate intermediate hidden representations \( h \), while the decoder $\mathcal{D}$ uses these representations to output the future motions.

In the context of continual learning, updating the parameters of an already trained neural network model is necessary to adapt to new domains. This introduces a crucial question: how can we ensure the model still performs well on previously learned domains? Techniques like EWC aim to protect those parameters that are critical for old domains. In contrast, some parameter isolation methods introduce a completely new set of parameters for training new domain-specific models. Similarly, our approach seeks to preserve previously trained parameters as much as possible, modifying only those necessary to accommodate the new domain. To achieve this, we maintain the encoder’s parameters unchanged during continual learning and only adjust the decoder’s parameters to develop a new specialized model for different domains. This requires access to a pre-trained generalized encoder with fixed parameters $\phi$ to produce consistent, universal hidden representations. As illustrated in in Fig. \ref{fig:framework}, the input $x_{[n]}$, where the subscript $[n]$ denotes the input from domain $n$, is encoded as:
\begin{equation}
\label{eq:encoder}
h_{[n]} = \mathcal{E}(x_{[n]};\phi).
\end{equation}
For the decoder, instead of introducing a new set of parameters for each domain, we employ a hypernetwork designed to learn and generate the required parameters $\theta^{(m)}$ for each specialized model $m$ specified by the superscript $(m)$, to predict future motions as:
\begin{equation}
\label{eq: decoder}
y \sim \mathcal{D}(h_{[n]};\theta^{(m)}).
\end{equation}

A hypernetwork $\mathcal{H}$ is a type of neural network that generates parameters for a target neural network. In our framework, we introduce a pool of domain queries $q^{(m)}$, which are high-dimensional vector representations of each specialized model. Each domain query serves as input to the hypernetwork, prompting it to generate specific parameters for each specialized decoder as:
\begin{equation}
\label{eq:hyper decoder}
\theta^{(m)} = \mathcal{H}(q^{(m)}; \Theta),
\end{equation}
where $\Theta$ denotes the parameters of the hypernetwork.
This approach avoids the storage burden associated with adding new model parameters in successive rounds of continual learning; instead, only new domain queries are added with each domain expansion. To mitigate catastrophic forgetting, we employ a hypernetwork-based continual learning approach, first introduced by Oswald et al. (2020) \cite{Oswald2020Continual}. This method ensures that while learning a new specialized model, the hypernetwork imposes constraints on the parameters generated for previous models. During the training of the $(m)$-th domain, the hypernetwork parameters $\Theta$ and current domain query $q^{(m)}$ are trained to minimize the following regularized loss function:
\begin{align}
\label{eq:total loss}
L_{\texttt{total}} &= L_{\texttt{motion}}(\Theta, q^{(m)}, X_{[m]}, Y_{[m]}) \notag \\
&\quad + L_{\texttt{domain}}(\Theta,q^{(m)},X_{[m]}) \notag \\
&\quad + L_{\texttt{reg}}(\Theta^*, \Theta, \Delta\Theta, \{q^{(m-1)}\}),
\end{align}
where $L_{\texttt{motion}}$ is the motion prediction loss on domain dataset $(X_{[m]}, Y_{[m]})$, $L_{\texttt{domain}}$ is a domain awareness loss which will be explained in Section \ref{sec:nf}, and the regularization term $L_{\texttt{reg}}$ is defined as:
\begin{equation}
\label{eq:reg loss}
L_{\texttt{reg}} = \lambda \sum_{i=1}^{m-1}||\mathcal{H}(q^{(i)}; \Theta^*)-\mathcal{H}(q^{(i)}; \Theta+\Delta\Theta)||^2.
\end{equation}
During continual domain expansion, this regularization term ensures that each previous domain query consistently generates the same parameters, with $\Theta^*$ denoting the hypernetwork parameters at the end of the last domain expansion.

\subsection{Normalizing Flow for Domain Awareness} \label{sec:nf}
Throughout Section \ref{sec:hyper}, we use both superscripts and subscripts to denote different specialized models and domains, respectively. This notation is essential as, in practice, domains do not have explicit labels. While we can align the current domain with a new specialized model during training as demonstrated in (\ref{eq:total loss}), during inference there are no explicit labels available at runtime to indicate which specialized model should be utilized.

Ideally, for any given scenario, we aim to deploy the specialized model that performs best based on certain metrics. The most straightforward metrics are displacement errors. However, these metrics require access to ground truth future trajectories, which are not available during inference. Therefore, our approach is predicated on the assumption that a specialized model's familiarity with a scenario correlates positively with its effectiveness in handling that scenario. More formally, considering the domain dataset used to train each specialized model, we attempt to model the distribution of the hidden representations, denoted as $\mathbb{P}_h$, of the training data. This allows us to estimate the likelihood that a given scenario fits within the trained model’s familiar contexts.

To accurately model such distributions, we propose using a normalizing flow \cite{rezende2015variational}, a versatile generative model well-suited for learning distributions of any shape directly. Normalizing flow effectively transforms the distribution of hidden representations, $\mathbb{P}_h$, into a standard Gaussian distribution, $\mathbb{P}_z$, via an invertible transformation $z=f(h;\omega)$. This transformation is expressed mathematically as:
\begin{equation}
\label{eq:nf loss}
\mathbb{P}_h(h;\omega) = \mathbb{P}_z(f(z;\omega)) \left| \det\frac{\partial f}{\partial h}\right|,
\end{equation}
where $\left| \det\frac{\partial f}{\partial h}\right|$ is the Jacobian determinant that adjusts for changes in volume induced by the transformation, ensuring the model captures the intricacies of the data distribution.

Specifically, we implement a normalizing flow using a series of affine coupling layers, a technique employed by several previous works \cite{dinh2014nice, kingma2018glow}, denoted as $f = f^1 \circ f^2 \circ \ldots \circ f^K$. With the input split into two subsets as $[h_a, h_b] = h$, each coupling layer executes an invertible transformation:
\begin{equation}
\label{eq:coupling layer}
(h'_a, h'_b) = f^k(h_a, h_b),
\end{equation}
where the transformation within a coupling layer is driven by a scale function $f_s^k$ and a shift function $f_t^k$, both parameterized by separate neural networks. The transformation equations are as follows:
\begin{align}
\label{eq:coupling layer2}
h'_a &= h_a \\
h'_b &= f_s^k(h_a;\omega^k_{s}) \odot h_b + f_t^k(h_a;\omega^k_{t}).
\end{align}
After transformation, the two subsets are concatenated to form the output $ h'=\texttt{concat}([h'_a,h'_b])$. To ensure full transformation of the input across the flow, the roles of the subsets are alternated in successive layers.

The design choice of learning the distribution of $h$ instead of directly over $x$ is motivated by several factors. Firstly, drawing inspiration from prior work \cite{wiederer2023joint} that utilized hidden features for a similar purpose, the feature representation provides a highly abstract depiction of the input scenario, which typically includes complex and heterogeneous contexts of vehicle trajectories and map data. Secondly, these features focus on prediction-critical information, enhancing model accuracy. Lastly, based on recommendations from previous literature \cite{kirichenko2020normalizing}, training normalizing flows directly from raw inputs poses challenges due to the requirement for invertible transformations, which complicates model architecture. Using encoded features simplifies the input complexity, thereby facilitating more manageable and effective model training. 

Given that normalizing flow operates as a neural network model, we seamlessly extend the hypernetwork to also provide parameters for it across different domains so that (\ref{eq:hyper decoder}) is expanded as:
\begin{align}
\label{eq: hyper nf}
\omega^{(m)} &= \mathcal{H}_\omega(q^{(m)}; \Theta_\omega), \notag \\
\theta^{(m)} &= \mathcal{H}_\theta(q^{(m)}; \Theta_\theta)
\end{align}

This integration allows us to maintain a consistent model architecture without needing to expand model parameters for each domain-specific flow model.

The training objective for normalizing flow involves minimizing the negative log-likelihood of the training data, expressed as: 
\begin{equation}
\label{eq: nf loss}
L_{\texttt{domain}} = -log \mathbb{P}_h(h_{[m]};\omega^{(m)}).
\end{equation}
During inference, the likelihood $P_h$ associated with each specialized model is computed to determine its familiarity with the current scenario. The specialized model chosen for deployment is the one that maximizes the likelihood of the current input under its domain-specific parameters, as follows:
\begin{equation}
\label{eq: evidence}
m^* = \underset {m} {\text{argmax}}\mathbb{P}_h(h;\omega^{(m)}).
\end{equation}
By gaining domain awareness, this process ensures that the model most accustomed to the nuances of the current scenario is selected, leveraging its domain-specific training to provide the most accurate predictions.

\subsection{Deep Bayesian Uncertainty Estimation} \label{sec:evid}
Continual domain expansion is triggered when incoming domain datasets lack a specialized model capable of delivering satisfactory performance during training. A similar challenge can arise during inference if none of the currently expanded specialized models perform effectively. For the task of motion prediction, this situation is particularly common. Thanks to the domain awareness provided by the normalizing flow, we can effectively detect such situations when all flow models output low likelihoods, indicating an out-of-distribution scenario. Since multi-modal motion prediction models must produce a set of future trajectories that jointly capture diverse motion intentions and sufficiently cover likely behaviors, selection mechanisms must consider both spatial diversity and the associated uncertainty of each trajectory. To this end, we propose to incorporate the uncertainty of the trajectories from the specialized models into the selection process, allowing us to identify underperforming predictions and fall back on generalist outputs when needed. The final component of our proposed DECODE framework introduces a performance lower bound, utilizing the generalized model to ensure coverage in these scenarios.

To do so, we incorporate deep Bayesian uncertainty estimation based on Natural Posterior Networks \cite{charpentier2022natural}. We begin by modeling the future motion $y$ using an exponential family distribution:
\begin{equation}
\label{eq: exponential family}
\mathbb{P}(y|\eta)=\beta(y)\exp{(\eta^TT(y)-A(\eta))},
\end{equation}
where $\eta$ denotes the natural parameter, $b(y)$ the base measure, $T(y)$ the sufficient statistic, and $A(\eta)$ the log-partition function. Given a domain dataset $D$, the posterior distribution is then updated using the Bayesian rule:
\begin{equation}
\label{eq: bayesian rule}
\mathbb{Q}(\eta|D) \propto \prod_{y\in D} \mathbb{P}(y|\eta) \times \mathbb{Q}(\eta).
\end{equation}
Given that the conjugate prior distribution $\mathbb{Q}(\eta)$ belongs to the same exponential family, the posterior can be parameterized as:
\begin{equation}
\label{eq: posterior}
\mathbb{Q}(\eta|\chi^{\tt{post}}, e^{\tt{post}}) \propto \exp{(e^{\tt{post}}\eta^T\chi^{\tt{post}}-e^{\tt{post}}A(\eta))},
\end{equation}
The updated posterior parameters are given by $\chi^{\tt{post}}=\frac{e^{\tt{prior}}\chi^{\tt{prior}}+\sum_{y\in D}T(y)}{e^{\tt{prior}}+|D|}$ and $e^{\tt{post}}=e^{\tt{prior}}+|D|$, with $\chi^{\tt{prior}}$ and $e^{\tt{prior}}$ denoting the prior parameters and prior evidence, respectively, and $|D|$ the size of the domain dataset. This formulation yields a weighted combination of the prior and the observed sufficient statistics, effectively balancing prior knowledge with new evidence.

The above formulation provides the foundation for defining a performance lower bound in our framework, achieved by using a pre-trained generalized model to predict the prior parameters. The decoder in~(\ref{eq: decoder}) is adapted to output these predicted $\chi$ values. Given $m^*$ as the ID of the best specialized model from~(\ref{eq: evidence}), the posterior parameters are then computed by combining predictions from both the generalized and specialized models. The integration is defined as:
\begin{align}
\label{eq: posterior update}
\chi^{\tt{post}}=\frac{e^{(0)}\chi^{(0)}+e^{(m^*)}\chi^{(m^*)}}{e^{(0)}+e^{(m^*)}}, e^{\tt{post}}=e^{(0)}+e^{(m^*)}.
\end{align}
Here, $\chi^{(0)}$ and $\chi^{(m^*)}$ represent the predicted parameters from the generalized and specialized models, respectively. A default constant value, generally small, is assigned to $e^{(0)}$ to represent the baseline evidence level. The domain-specific evidence is then estimated using the normalizing flow, as described in the following equation:
\begin{equation}
\label{eq: evidence nf}
e^{(m^*)} = \mathbb{P}_h(h;\omega^{(m^*)}).
\end{equation}

\begin{algorithm}
\caption{Sampling general DECODE framework}
\label{alg:sampling}
\begin{algorithmic}[1] 
\Statex \textbf{Input:} Domain queries $\{q^{(i)}\}_{i=1}^m$, scenario input $x$, number of sample $K$.
\Statex \textbf{Returns:} A set of future motion samples $\{ \hat{y}^{k} \}_{k=1}^K$.
\State \textbf{Encode:}
\State \hskip\algorithmicindent Compute the hidden representation $h = \mathcal{E}(x;\phi)$.
\State \hskip\algorithmicindent Find the best specialized model 
\Statex \hskip\algorithmicindent $e^{(m^*)} = \underset {m} {\tt{max}}\mathbb{P}_h(h;\mathcal{H}_\omega(q^{(m)}; \Theta_\omega))$
\State \textbf{Decode:}
\State \hskip\algorithmicindent Compute parameters from the best specialized model 
\Statex \hskip\algorithmicindent $\chi^{(m^*)} = \mathcal{D}(h;\mathcal{H}_\theta(q^{(m^*)}; \Theta_\theta))$
\State \hskip\algorithmicindent Compute parameters from the generalized model 
\Statex \hskip\algorithmicindent $\chi^{(0)} = \mathcal{D}(h;\theta^{(0)})$
\State \hskip\algorithmicindent Compute posterior parameters
\Statex \hskip\algorithmicindent $\chi^{\tt{post}}=\frac{e^{(0)}\chi^{(0)}+e^{(m^*)}\chi^{(m^*)}}{e^{(0)}+e^{(m^*)}}, e^{\tt{post}}=e^{(0)}+e^{(m^*)}$
\Statex 
\For{$k = 1$ \textbf{to} $K$}
\State \hskip\algorithmicindent Sample parameters $\eta^k \sim \mathbb{Q}(\eta|\chi^{\tt{post}}, e^{\tt{post}})$ \Comment{Alternatively compute the mean parameter $\Bar{\eta}$ for deterministic sampling}
\State \hskip\algorithmicindent Sample future motion $\hat{y}^k \sim \mathbb{P}(y|\eta^k)$. 
\EndFor
\State
\State \textbf{return} $\{ \hat{y}^{k} \}_{k=1}^K$.
\end{algorithmic}
\end{algorithm}

Following the Bayesian update mechanism, we have developed a practical algorithm, summarized in Algorithm \ref{alg:sampling}. As depicted in Algorithm \ref{alg:sampling}, when the evidence supporting the specialized model exceeds that of the generalized model, the posterior parameters $\chi^{\tt{post}}$ will align more closely with $\chi^{(m^*)}$. Conversely, if the evidence favors the generalized model, the posterior parameters $\chi^{\tt{post}}$ will resemble $\chi^{(0)}$ more closely. Consequently, this ensures that the performance is always at least as good as that provided by the generalized model. Regarding the training objective, a Bayesian loss for the $(m)$-th domain expansion is articulated as follows:
\begin{equation}
\label{eq: bayesian loss}
L_{\texttt{motion}} = -\mathbb{E}_{\eta_{[m]}\sim\mathbb{Q}^{\tt{post},(m)}}[\log \mathbb{P}(y_{[m]}|\eta_{[m]})]-\mathbb{H}[\mathbb{Q}^{\tt{post},(m)}].
\end{equation}

\section{Model Instantiation with Motion Transformer}
\subsection{Motion Prediction Model Implementation} \label{sec: MTR inplem}
Our proposed continual learning framework is designed to be compatible with any motion prediction model as long as it follows an encoder-decoder structure. For illustrative purposes, we have chosen the Motion Transformer (MTR) model \cite{shi2022mtr, shi2024mtr++}, a state-of-the-art, multi-modal motion prediction algorithm, to demonstrate the practical implementation of our framework. The MTR model utilizes a transformer-based encoder-decoder architecture, enhancing motion prediction in autonomous driving scenarios. It features learnable intention queries that efficiently and accurately predict future trajectories across various motion modalities, eliminating the need for a dense set of goal candidates. This model combines global intention localization, which identifies an agent’s future intentions to improve prediction efficiency, with local movement refinement for greater accuracy in trajectory adjustments. Demonstrating superior performance in competitive benchmarks, the MTR model exemplifies the effectiveness of incorporating advanced motion prediction models within our framework. Readers interested in a deeper exploration of the MTR model are encouraged to consult the original work \cite{shi2022mtr, shi2024mtr++}.

\begin{figure}[!t]
\centering
\includegraphics[width=3.4in]{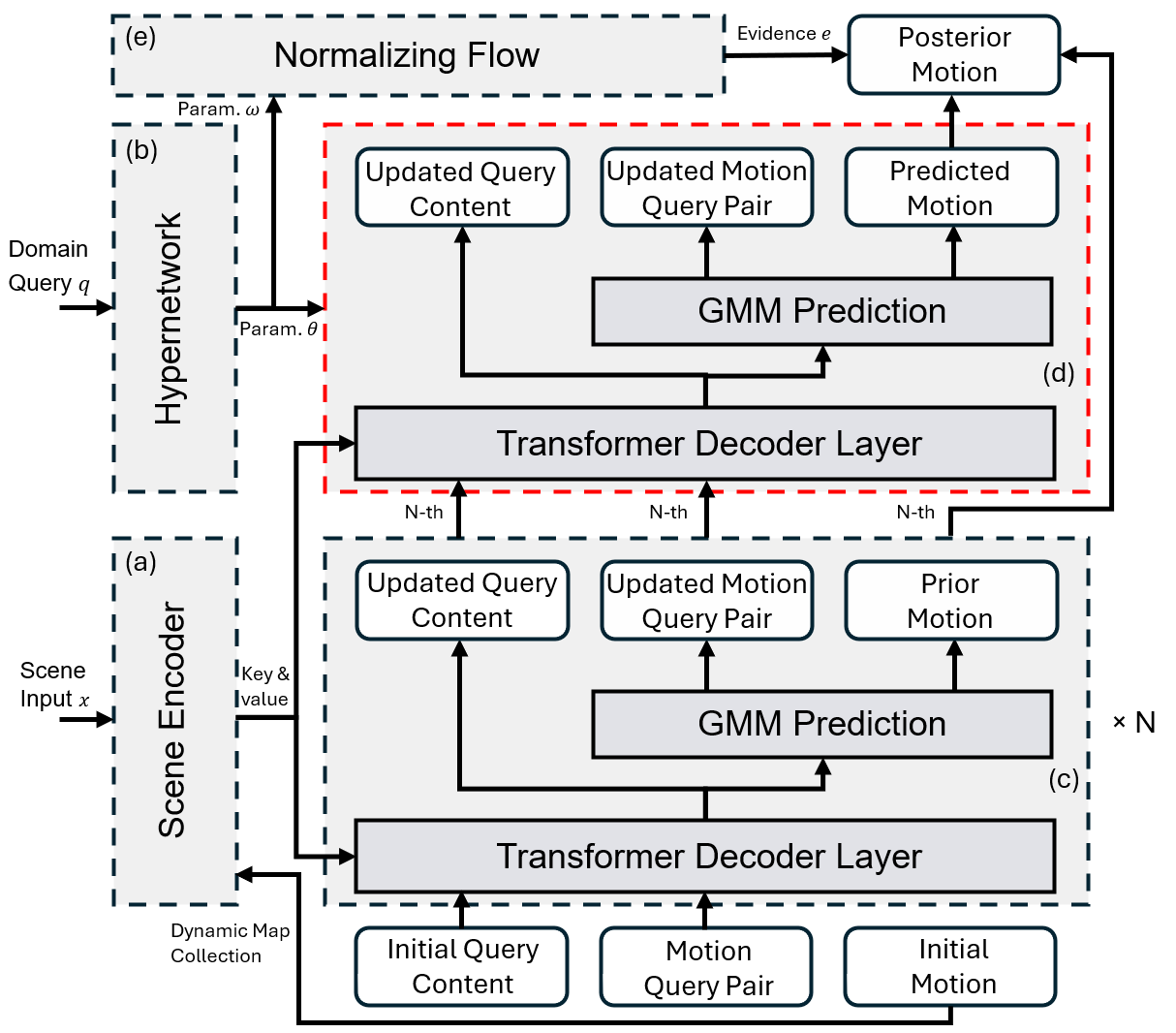}
\caption{Modified Architecture of MTR: (a) Scene Encoder processes scene input $x$ and outputs hidden representations for vehicles and lanes. (b) Hypernetwork receives a domain query $q$ and generates parameters for (d) the specialized decoder and (e) the specialized normalizing flow. (c) Generalized decoder, like (a), is pre-trained and fixed to ensure stability. Details of (a) and (b) are omitted to focus on the decoding process. During decoding, outputs from (c) feed into (d), functioning as an additional layer in the transformer decoder. With the evidence $e$ from (e), and the parameters $\chi_{[0]}$ from (c) and $\chi_{[m^*]}$ from (d), the posterior parameters are updated, ultimately leading to the sampling of the predicted future motion.}
\label{fig_3}
\end{figure}

The original MTR model, with over 60 million trainable parameters, is significantly larger than other state-of-the-art motion prediction models. This extensive parameter size not only poses challenges for integrating with our DECODE framework, where decoder parameters must be generated by a hypernetwork, but also complicates stable and effective learning due to the complexities involved in managing such a large neural network. Despite these challenges, the DECODE framework is designed with the flexibility to accommodate and adapt to these conditions. Given that the MTR’s decoder consists of several motion transformer layers that share the same structure, input, and output characteristics, our implementation strategically adds an additional motion transformer layer at the end of the original decoder. The hypernetwork dynamically supplies the parameters for this new layer for each specialized model while the rest of the decoder remains fixed and acts as the generalized model, as depicted in Fig. \ref{fig_3}.

The original MTR model outputs future motions as a distribution represented by a Gaussian Mixture Model (GMM). Following \ref{eq: exponential family} in the last section, the posterior distribution can be similarly modeled as a GMM. Given that a GMM combines component selection, represented as a categorical distribution, with trajectory generation for each component modeled as a multivariate Gaussian distribution, we have simplified our approach for practicality. During training, we found it sufficient to perform the Bayesian update only for the component selection part, while maintaining the original settings for trajectory generation. For the categorical distribution, the conjugate prior is formulated as a Dirichlet distribution $\eta \sim \tt{Dir}(\alpha)$, where $\alpha = e \chi$. Regarding trajectory sampling, commonly only a limited number of future trajectories are generated to represent the distribution of future motion. Thus, we combine the components of the generalized and specialized models, and employ non-maximum suppression (NMS) to select the components along with their corresponding predicted trajectories.

\subsection{Hypernetwork Model Implementation}
Hypernetworks, while offering substantial capabilities, pose significant training challenges due to their role in generating parameters for primary neural networks, particularly within the complex landscape of motion prediction. These challenges are further amplified in state-of-the-art motion prediction models that employ intricate neural network structures, such as transformers. To address these issues, our framework incorporates a variety of techniques designed to facilitate the training of hypernetworks, ensuring efficient and stable learning processes.

\begin{figure}[!t]
\centering
\includegraphics[width=3in]{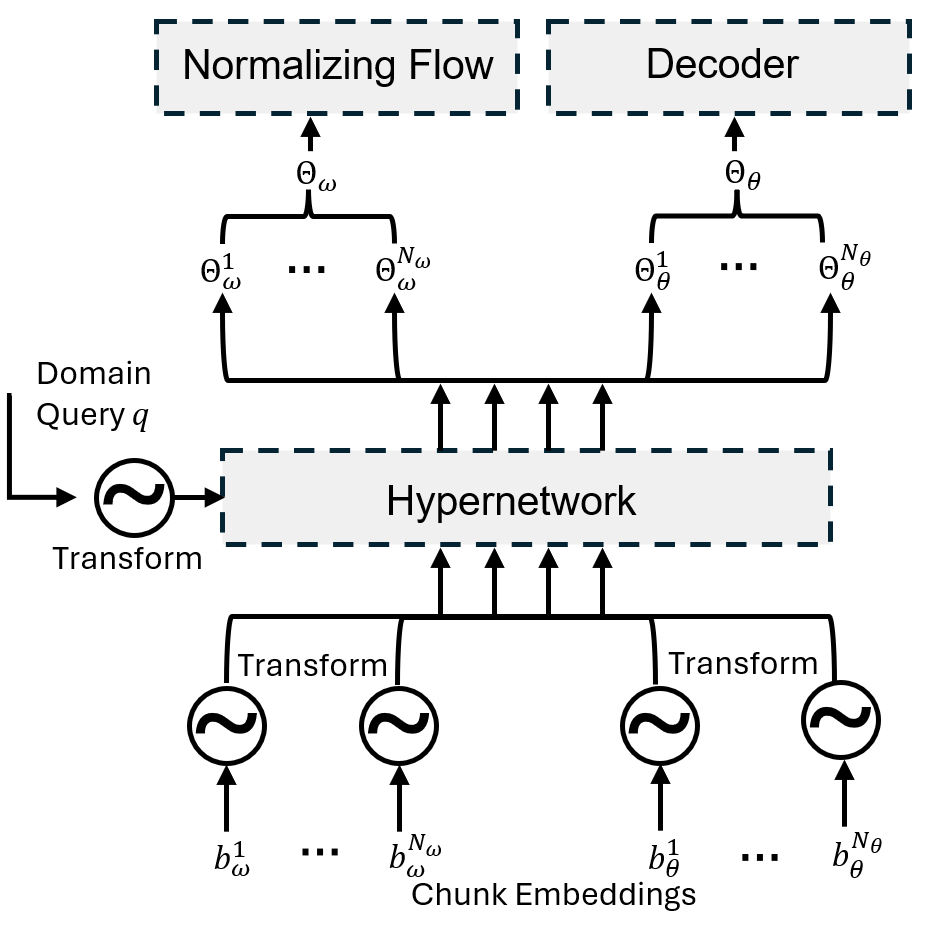}
\caption{Structure of the chunked hypernetwork: The hypernetwork is designed to reuse its model structure by incorporating two distinct sets of chunk embeddings, 
$\{b_\omega^i\}_{i=1}^{N_\omega}$ for the normalizing flow and  $\{b_\theta^i\}_{i=1}^{N_\theta}$ for the decoder. Each domain query 
$q$ is transformed and then concatenated with each set of chunk embeddings before being fed into the hypernetwork. The resulting outputs are 
$\Theta_\omega=\{\Theta_\omega^i\}_{i=1}^{N_\omega}$
  and $\Theta_\omega=\{\Theta_\theta^i\}_{i=1}^{N_\theta}$, which respectively form the parameters for the normalizing flow and the decoder, as depicted in the arrows leading from the outputs of the hypernetwork to their respective application modules.
 }
\label{fig_4}
\end{figure}

First, following the strategy outlined in the original work on continual learning with hypernetworks, we have adopted 'chunking' to reduce the complexity of generating parameters for the target neural networks, thereby diminishing the hypernetwork's size. As depicted in Fig. \ref{fig_4}, chunking introduces two sets of chunk embeddings $\{b_\omega^i\}_{i=1}^{N_\omega}$ and $\{b_\theta^i\}_{i=1}^{N_\theta}$ as additional inputs to the hypernetwork. These embeddings, which are parts of the hypernetwork model, are concatenated with the domain query $q$ to serve as inputs, with each set producing a distinct segment of parameters for the target neural network. This method allows for efficient reuse of the hypernetwork, streamlining the parameter generation process for complex architectures.

Secondly, to enhance the stability of hypernetwork training, we utilize magnitude-invariant parametrizations, as introduced in \cite{ortiz2024magnitude}. These parametrizations transform domain queries into a space where each has a constant Euclidean norm of 1. This is achieved through a transformation function that combines cosine and sine operations, ensuring that the magnitude of inputs to the hypernetwork remains consistent as is shown in Fig. \ref{fig_4}.

Lastly, to address the issue of parameter explosion during training—a common complication with parameters generated by the hypernetwork—we have implemented a principled weight initialization method as described in \cite{Chang2020Principled}. This method carefully scales the initial weights of the hypernetwork to ensure they are appropriately sized. Such meticulous initialization not only stabilizes the network throughout the training process but also enhances convergence. Effectively managing the growth of parameters, this approach not only avoids computational inefficiencies but also bolsters the model’s overall performance and reliability in complex motion prediction scenarios.
\section{Experiments}
\subsection{Experimental Setup}
\textbf{Domain datasets:} To evaluate the performance of our proposed DECODE framework, we employ several domain-specific datasets essential to motion prediction studies, simulating the continuous data flow characteristic of continual learning environments. Specifically, we employ the RounD dataset \cite{rounDdataset}, denoted as $\tt{D1}$, which consists of 76,112 training scenes and 12,416 validation scenes in roundabout scenarios; the HighD dataset \cite{highDdataset}, denoted as $\tt{D2}$, with 91,160 training scenes and 28,032 validation scenes in highway scenarios; and the InD dataset \cite{highDdataset}, denoted as $\tt{D3}$, which includes 8,056 training scenes and 2,280 validation scenes in urban intersections. These varied settings help test our framework’s robustness across different traffic environments. In each settings, the past observation spans 1 second, while the prediction horizon extends to 6 seconds into the future, with a timestep of 0.1 second. 
\begin{table*}[!b]
\caption{Performance comparison of motion prediction in different learning phases\label{tab:table1}}
\centering
\begin{tabular}{c|c|c|c|c|c|c}
\toprule[2pt]
\hline
Method & $\tt{minADE}_{\tt{D1, \_}}\downarrow$ & $\tt{minFDE}_{\tt{D1, \_}}\downarrow$ & $\tt{minADE}_{\tt{D2, \_}}\downarrow$  & $\tt{minFDE}_{\tt{D2, \_}}\downarrow$  &  $\tt{minADE}_{\tt{D3, \_}}\downarrow$  & $\tt{minFDE}_{\tt{D3, \_}}\downarrow$ \\
& ($\tt{P1}\rightarrow \tt{P2}\rightarrow \tt{P3}$) & ($\tt{P1}\rightarrow \tt{P2}\rightarrow \tt{P3}$) & ($\tt{P2}\rightarrow \tt{P3}$) & ($\tt{P2}\rightarrow \tt{P3}$) & ($\tt{P3})$ & ($\tt{P3}$) \\
\hline
$\texttt{oriMTR}$ & $\textbf{0.470}\rightarrow9.008\rightarrow2.487$ & $\textbf{1.153}\rightarrow21.639\rightarrow6.668$ & $0.442\rightarrow23.505$ &$1.202\rightarrow53.436$ &$0.758$ & $1.987$ \\
$\texttt{preMTR}$ & $0.938\rightarrow0.938\rightarrow0.938$ & $2.292\rightarrow2.292\rightarrow2.292$ & $0.906\rightarrow0.906$ & $2.320\rightarrow2.320$ & $1.127$ & $3.054$  \\
$\texttt{EWC}$ &$0.520\rightarrow2.345\rightarrow1.819$ & $1.224\rightarrow6.276\rightarrow5.025$ & $\textbf{0.364}\rightarrow6.429$ & $\textbf{0.960}\rightarrow 16.827$& $0.755$& $2.030$ \\
$\texttt{MAS}$ & $0.517\rightarrow3.330\rightarrow1.759$ & $1.240\rightarrow8.300\rightarrow4.436$ & $0.395\rightarrow5.120$ & $1.059\rightarrow13.18$ & $0.768$ & 
$2.030$  \\
$\texttt{ER(500)}$ & $0.470\rightarrow1.680\rightarrow1.313$ & $1.153\rightarrow4.112\rightarrow3.146$ & $0.443\rightarrow0.778$ & $1.121\rightarrow1.751$ & $0.712$ & 
$1.889$  \\
$\texttt{DER++(500)}$ & $0.470\rightarrow0.781\rightarrow0.713$ & $1.153\rightarrow1.764\rightarrow1.669$ & $0.421\rightarrow0.492$ & $1.096\rightarrow1.354$ & $0.711$ & 
$1.929$  \\
$\texttt{Dualprompt}$ & $0.547\rightarrow0.591\rightarrow0.618$ & $1.290\rightarrow1.381\rightarrow1.398$ & $0.447\rightarrow\textbf{0.452}$ & $1.217\rightarrow\textbf{1.229}$ & $0.996$ & 
$2.541$  \\
$\texttt{Coda-P}$ & $ 0.494\rightarrow0.710\rightarrow0.976$ & $1.173\rightarrow1.632\rightarrow2.263$ & $0.371\rightarrow0.899$ & $1.060\rightarrow2.488$ & $\textbf{0.648}$ & 
$\textbf{1.681}$  \\
$\texttt{DECODE}$ & $0.523\rightarrow\textbf{0.525}\rightarrow\textbf{0.578}$ & $1.262\rightarrow\textbf{1.268}\rightarrow\textbf{1.319}$ & $0.520\rightarrow0.595$ & $1.262\rightarrow 1.278$ & $0.765$ & $1.982$ \\
\hline
\bottomrule[2pt]
\end{tabular}
\end{table*}
\textbf{Learning phases:} Our multi-phase continual learning process is organized into three phases: $\tt{P1}$, $\tt{P2}$, and $\tt{P3}$, corresponding to the sequential learning of the $\tt{D1}$ (RounD), $\tt{D2}$ (HighD), and $\tt{D3}$ (InD) domains. This approach is structured so that the model sequentially trains on $\tt{D1}$,
$\tt{D2}$, and $\tt{D3}$ in a streaming fashion. In each phase, only the data relevant to the current domain are available for training, and validation sets from previous domains remain accessible for performance evaluations.

\textbf{Performance metrics:} To effectively evaluate the performance of continual learning within our framework, we compute established motion prediction metrics such as minimum average displacement error (minADE) and minimum final displacement error (minFDE). Additionally, in line with previous studies on continual learning, we include metrics specifically designed to assess continual learning performance \cite{ ma2021continual, yang2022continual}: average error rate (AER) and forgetting rate (FGT), as follows:
\begin{align}
\label{eq:aer fgt}
\tt{AER} &= \frac{1}{N(N+1)/2}\sum_{j\geq i}^N \tt{R}_{\tt{Di},\tt{Pj}}, \\
\tt{FGT} &= \frac{1}{N(N-1)/2}\sum_{i=1}^N\sum_{j > i}^N \tt{R}_{\tt{Di},\tt{Pj}}-\tt{R}_{\tt{Di},\tt{Pi}},
\end{align}
where $N$ is the number of domain datasets / learning phases, and $\tt{R}_{\tt{Di}, \tt{Pj}}$ denotes either minADE or minFDE for domain $i$ after learning phase $j$.
The average error rate (AER) quantifies the overall average performance based on minADE or minFDE at the end of the continual learning cycle, providing a benchmark of the model's accuracy across all phases. The forgetting rate (FGT) assesses the degree to which the model retains previously learned information throughout the learning process, with lower values indicating better preservation of knowledge. 

\textbf{Implementation details:} For the implementation of the MTR model, we adhered to the UniTraj framework \cite{feng2024unitraj} and trained the generalized model using the Waymo Open Motion Dataset (WOMD) \cite{chen2023womdlidar}, which comprises 487,000 training scenes. The map polylines for the three domain datasets are aligned with the WOMD definitions, where each polyline consists of 20 points covering approximately 10 meters, and 700 nearest polylines are considered relative to the target vehicle. Both the encoder and the decoder are composed of six transformer layers, with each layer having a hidden feature dimension of 256. As discussed in Section \ref{sec: MTR inplem}, only the parameters for the last transformer layer of the decoder are dynamically generated by the hypernetwork, while the rest are fixed. In the MTR settings, a set of 64 initial motion queries are generated using a k-means algorithm applied to data points from the WOMD dataset; these queries remain unchanged during the training on new domain datasets. Multi-modality is addressed by predicting six possible future outcomes. The normalizing flow component of our model includes eight coupling layers, each with a hidden dimension of 256. Code is available at: \href{https://github.com/michigan-traffic-lab/DECODE}{https://github.com/michigan-traffic-lab/DECODE}. 

The hypernetwork, responsible for generating parameters for both the decoders and normalizing flows, receives inputs from domain queries and chunk embeddings, each with a dimension of 128. The hypernetwork's architecture consists of multi-layer perceptrons (MLP) with a hidden layer configuration of [256, 256] for each sub-layer of the transformer decoder and the coupling layers. The output chunking dimensions are set to 16384 for the object cross-attention layer, 8192 for the map cross-attention layer, 1280 for each coupling layer, and 4096  for other MLP layers in the decoder, such as the motion head.

\begin{figure*}[!t]
      \centering
      \includegraphics[width=7.0in]{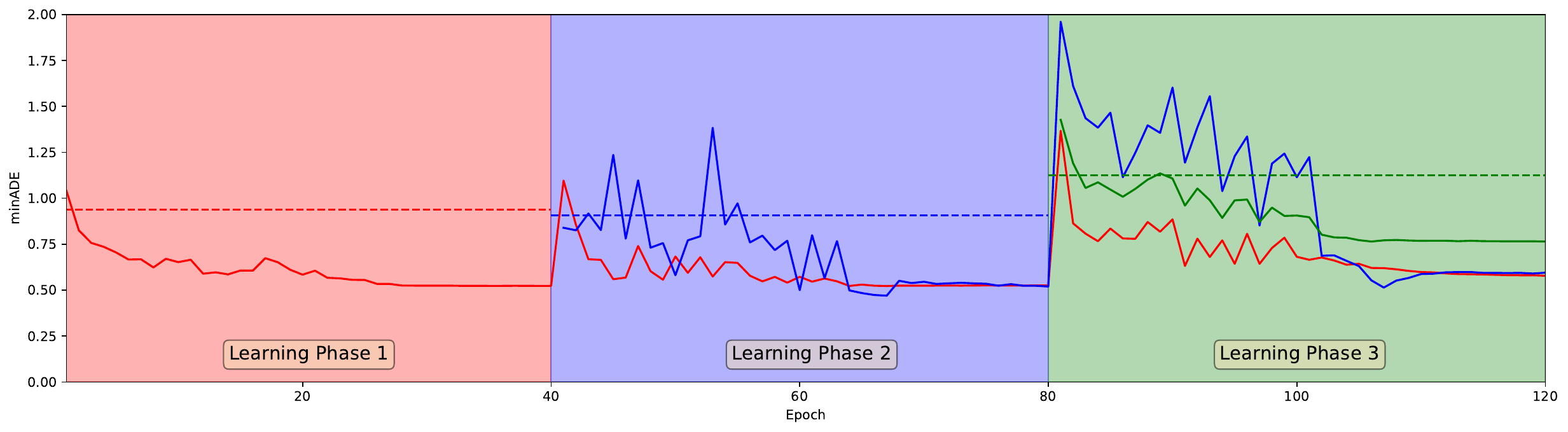}
      \caption{Epoch vs. minADE on Validation Sets in the DECODE Framework Across Three Learning Phases: Solid lines represent the performance of the proposed DECODE framework, while dashed lines indicate the benchmark performance of the pre-trained generalized model. The color coding is used to distinguish between domains: red for the RounD (D1) domain, blue for the HighD (D2) domain, and green for the InD (D3) domain.}
      \label{fig:main results}
\end{figure*}
\textbf{Training details:} The AdamW optimizer is utilized, chosen for its effectiveness in handling sparse gradients and improving generalization. We set the initial learning rate at $1e^{-4}$, adjusting it according to a predefined schedule where it is halved every 2 epochs after the 20 epochs to refine the learning as training progresses. The model was trained with a batch size of $128$ and a weight decay of $1e^{-2}$ to balance training speed and model performance. The training was conducted over 40 epochs using 2 H100 GPUs.
\begin{table}[!b]
\caption{Performance comparison of continual learning metrics\label{tab:table2}}
\centering
\setlength{\tabcolsep}{2.8pt}
\begin{adjustbox}{max width=\columnwidth}
\begin{tabular}{c|c|c}
\toprule[2pt]
\hline
Method & $\tt{AER}\downarrow$ & $\tt{FGT}\downarrow$ \\
& ($\tt{minADE}$ / $\tt{minFDE}$) & ($\tt{minADE}$ / $\tt{minFDE}$) \\
\hline
$\texttt{oriMTR}$ & $6.110\pm0.024$ / $14.335\pm0.052$ & $11.205\pm0.060$ / $26.057\pm0.120$  \\
$\texttt{preMTR}$ & $0.959\pm0.016$ / $2.426\pm0.039$ & $-$  \\
$\texttt{EWC}$ & $2.038\pm0.008$ / $5.390\pm0.022$ & $3.063\pm0.012$ / $8.240\pm0.034$  \\
$\texttt{MAS}$ & $1.982\pm0.009$ / $5.043\pm0.025$ & $2.929\pm0.020$ / $7.461\pm0.053$  \\
$\texttt{ER(500)}$ & $0.896\pm0.008$ / $2.193\pm0.026$ & $0.789\pm0.006$ / $1.860\pm0.021$  \\
$\texttt{DER++(500)}$ & $0.598\pm0.010$ / $1.494\pm0.023$ & $0.209\pm0.007$ / $0.466\pm0.017$  \\
$\texttt{Dualprompt}$ & $0.608\pm0.009$ / $1.509\pm0.037$ & $\mathbf{0.040\pm0.002}$ / $0.062\pm0.009$  \\
$\texttt{Coda-P}$ & $0.647\pm0.006$ / $1.633\pm0.018$ & $0.280\pm0.006$ / $0.698\pm0.017$  \\
$\texttt{DECODE}$ & $\mathbf{0.584\pm0.005}$ / $\mathbf{1.395\pm0.019}$ & $0.044\pm0.001$ / $\mathbf{0.027\pm0.001}$  \\
\hline
\bottomrule[2pt]
\end{tabular}
\end{adjustbox}
\end{table}

\subsection{Main Results}
Table \ref{tab:table1} presents the performance evaluation of our proposed DECODE framework by comparing performance metrics across multiple learning phases. For each column, we illustrate how performance evolves after each learning stage, denoted as $\tt{minADE}_{\tt{Di}, \tt{Pj}}$ and $\tt{minFDE}_{\tt{Di}, \tt{Pj}}$, where $j \geq i$ indicates the metric for domain $i$ after learning phase $j$. For consistency, all models utilize the same MTR backbone architecture. The comparison lineup includes: (1) the original MTR model ($\texttt{oriMTR}$)\cite{shi2022mtr}, trained from scratch on each current domain dataset without any continual learning strategy; (2) a pre-trained generalized MTR model ($\texttt{preMTR}$) trained on the whole WOMD dataset, used directly without further domain adaptation; (3) regularization-based continual learning baselines, including Elastic Weight Consolidation ($\texttt{EWC}$) \cite{kirkpatrick2017overcoming} and Memory Aware Synapses ($\texttt{MAS}$) \cite{aljundi2018memory}; (4) rehearsal-based baselines, including a standard Experience Replay ($\texttt{ER}$) buffer and Dark Experience Replay ($\texttt{DER++}$) \cite{buzzega2020dark}; and (5) state-of-the-art parameter-efficient tuning methods, namely DualPrompt \cite{wang2022dualprompt} and CODA-Prompt \cite{smith2023coda}, adapted to operate with the pre-trained MTR backbone.

The results highlight significant performance fluctuations in the $\texttt{oriMTR}$, which lacks any continual learning strategy. In contrast, the performance of the $\texttt{preMTR}$ remains static, as it does not engage in any continual learning. Notably, the generalized model’s performance, while stable, does not surpass that of $\texttt{oriMTR}$ when initially trained on a domain. A substantial increase in displacement errors is evident when transitioning from the RounD to the HighD domain, attributable to the distinct driving patterns of these domains. However, performance improves upon moving to the InD domain, reflecting the similarity in driving conditions between RounD and InD. Regularization-based methods like $\texttt{EWC}$ and $\texttt{MAS}$ perform notably worse than rehearsal-based approaches such as $\texttt{ER}$ and $\texttt{DER++}$, indicating that replaying past samples is more effective than parameter regularization for reducing forgetting in motion prediction. Noticeably, since the original $\texttt{DER++}$ framework applies soft-label matching on logits for pure classification tasks, in our motion prediction setting we adopt this strategy only for modality updates while using standard rehearsal for regression updates. For both $\texttt{DualPrompt}$ and $\texttt{CODA-P}$, we report results using the prompts inserted into both encoder and decoder, and our proposed method consistently outperforms or matches these approaches across all learning phases. Overall, the DECODE strategy demonstrates the most stable performance among all compared methods, exhibiting only a slight increase in errors during the initial phase of each new domain and maintaining relatively steady performance thereafter.

Table \ref{tab:table2} presents a comparison of the Average Error Rate (AER) and Forgetting Measure (FGT) for each strategy following the completion of the three-phase continual learning training. The results demonstrate that the proposed DECODE strategy outperforms all others in terms of AER, indicating its efficacy in adapting to new domains without significant performance loss. The pre-trained generalized model exhibits an effective forgetting rate of zero since it undergoes no further training; however, its AER is considerably higher than that of DECODE. This highlights the advantage of additional learning phases in developing specialized models that better accommodate new information. Notably, DECODE also shows the lowest forgetting rate among the strategies that involve active learning. The online EWC strategy underperforms, suggesting that the large number of parameters required for EWC may complicate effective learning and adaptation.
\begin{figure*}[!t]
      \centering
        \subfloat[D1-P1]{\includegraphics[trim=0 0 60 0, clip, height=1.45in]{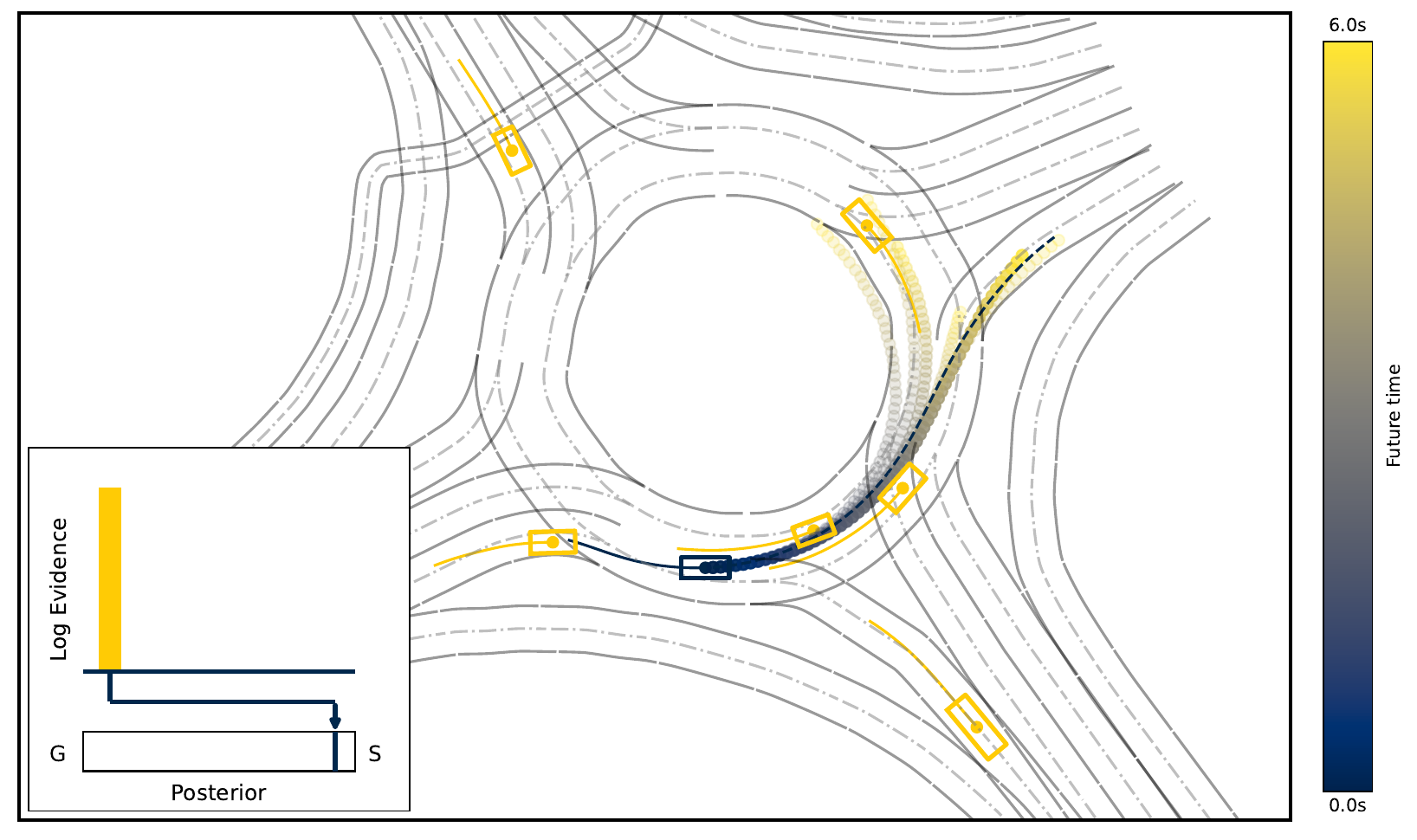}%
        \label{fig: visual a}}    
        \subfloat[D1-P2]{\includegraphics[trim=0 0 60 0, clip, height=1.45in]{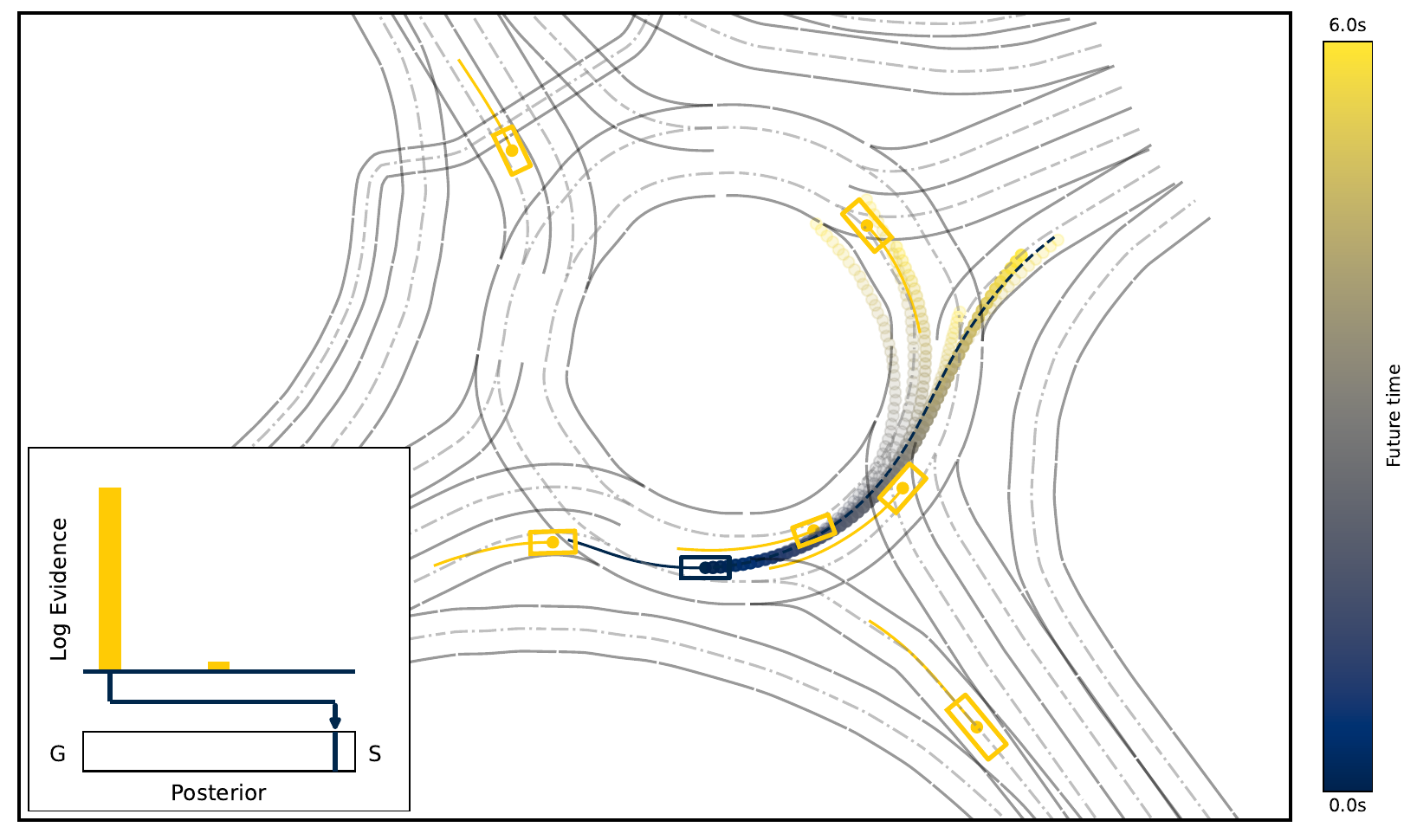}%
        \label{fig: visual b}}
        \subfloat[D1-P3]{\includegraphics[height=1.45in]{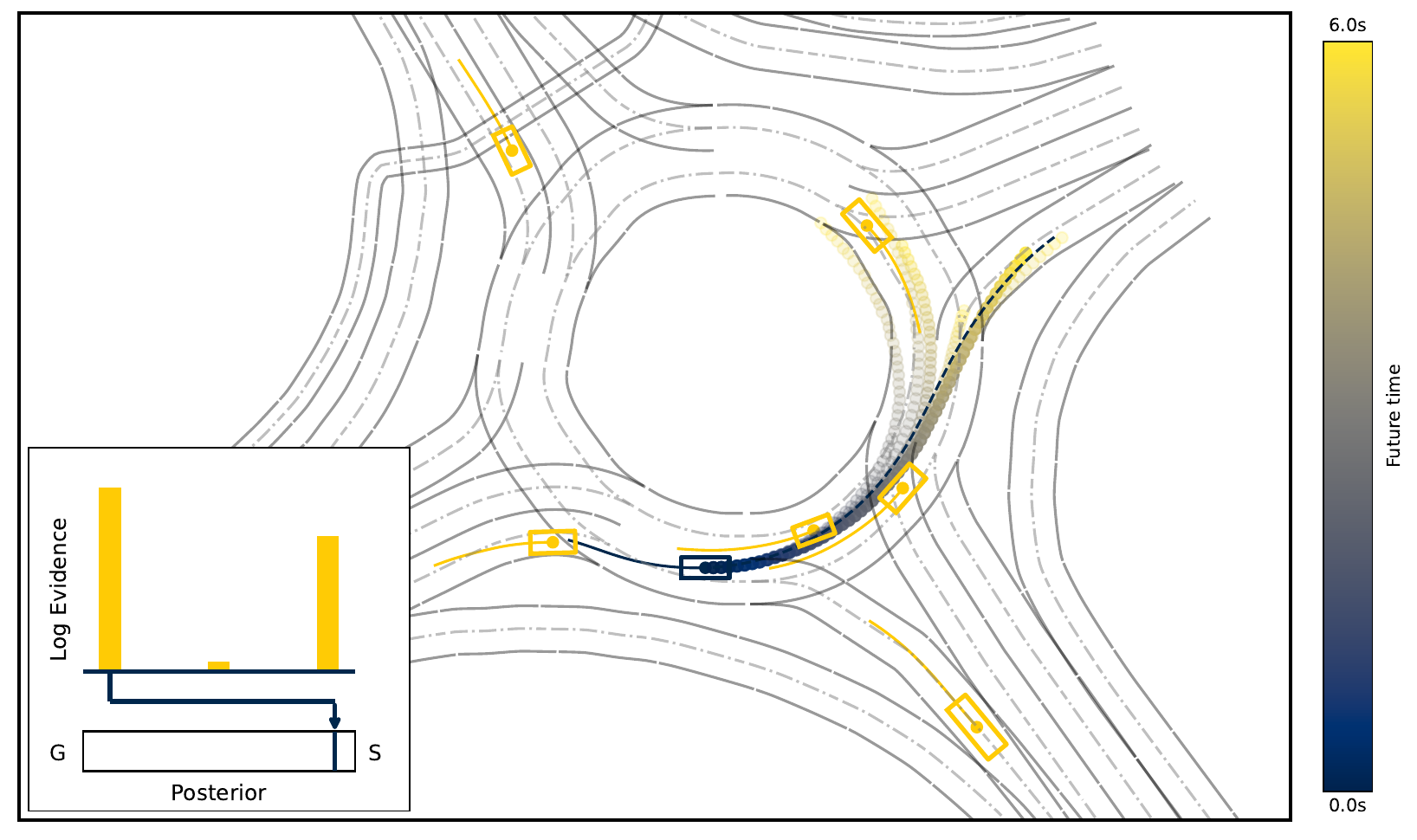}%
        \label{fig: visual c}} \\
        \subfloat[D2-P1]{\includegraphics[trim=0 0 60 0, clip, height=1.25in]{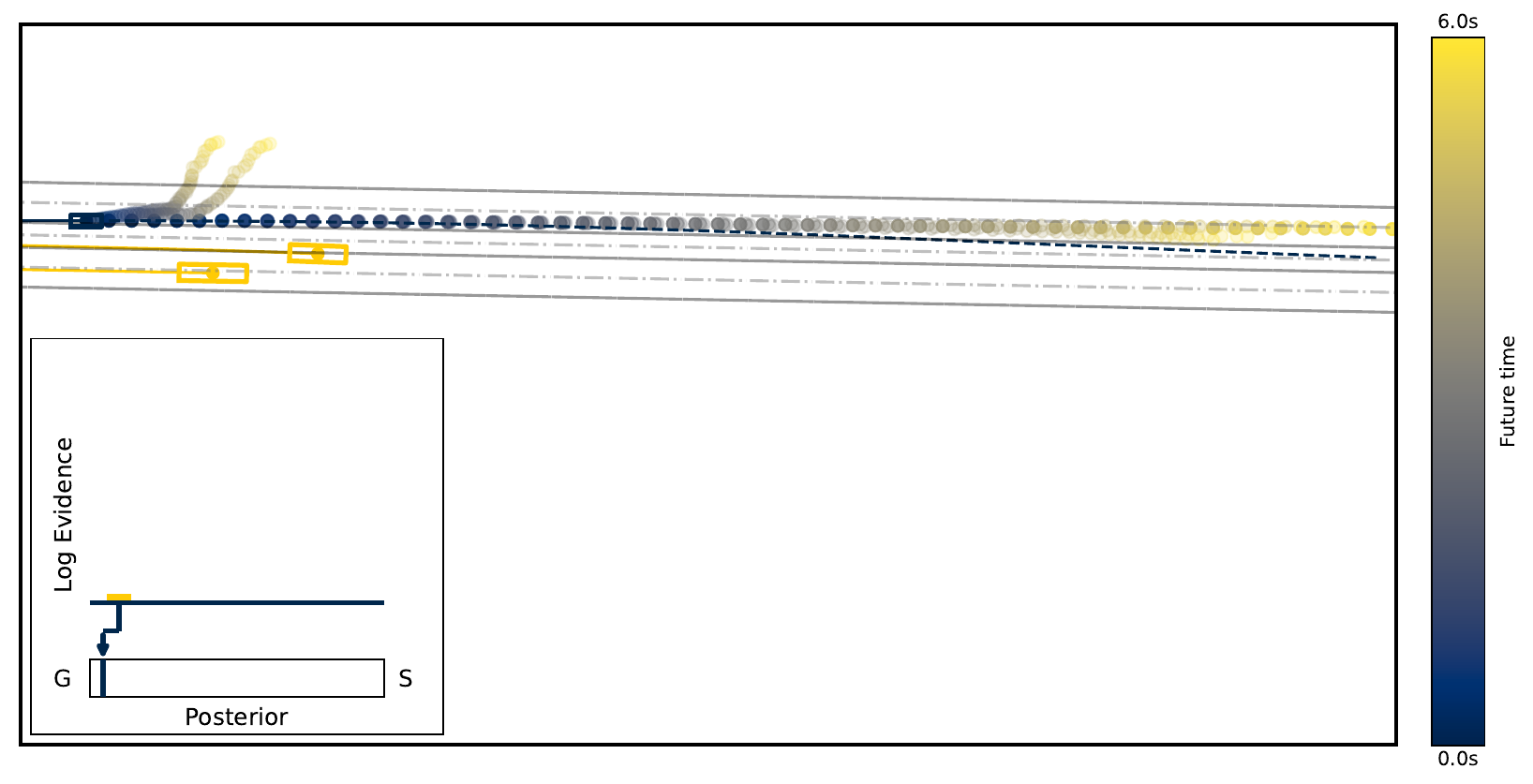}%
        \label{fig: visual d}}
        \subfloat[D2-P2]{\includegraphics[trim=0 0 60 0, clip, height=1.25in]{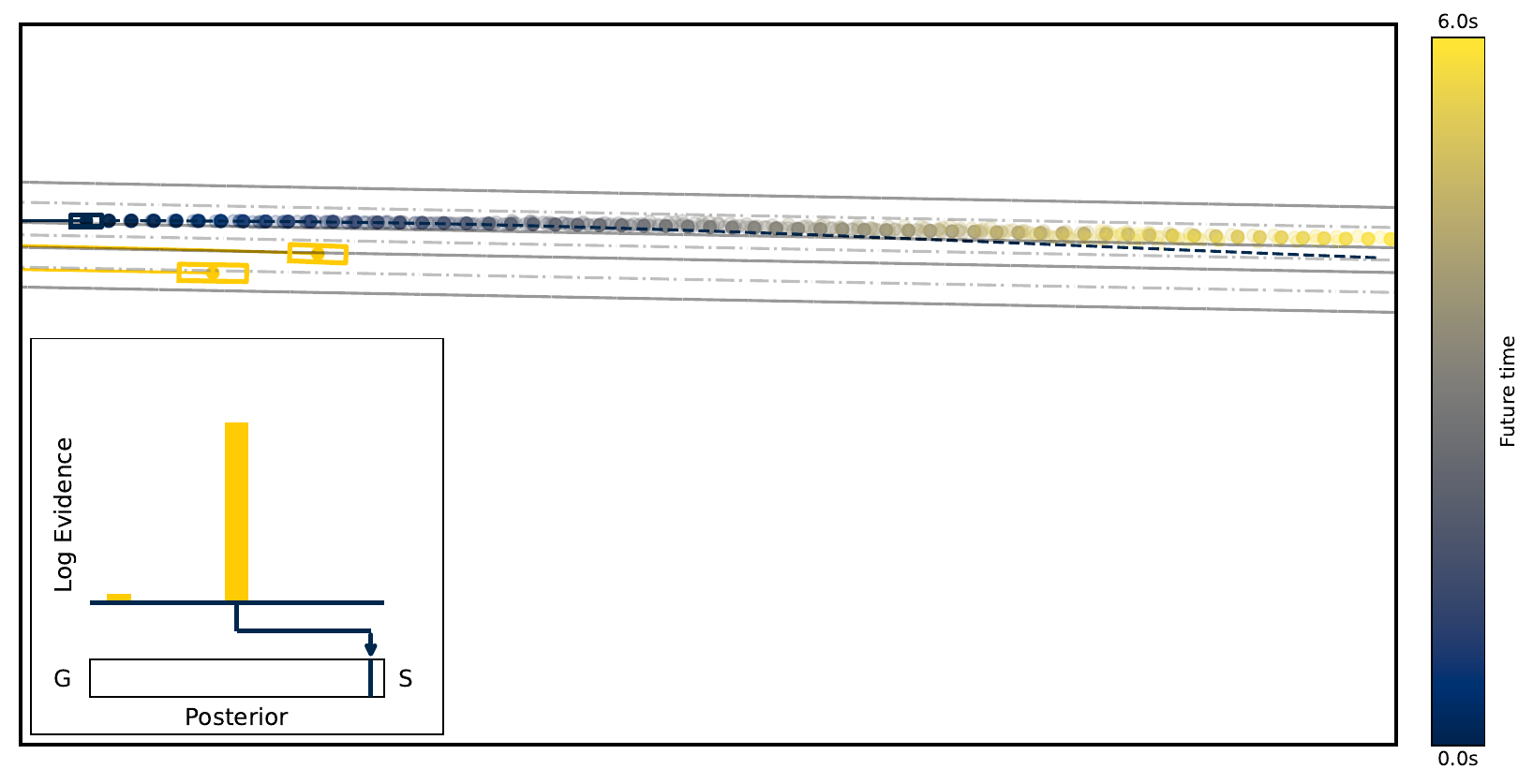}%
        \label{fig: visual e}}
        \subfloat[D2-P3]{\includegraphics[height=1.25in]{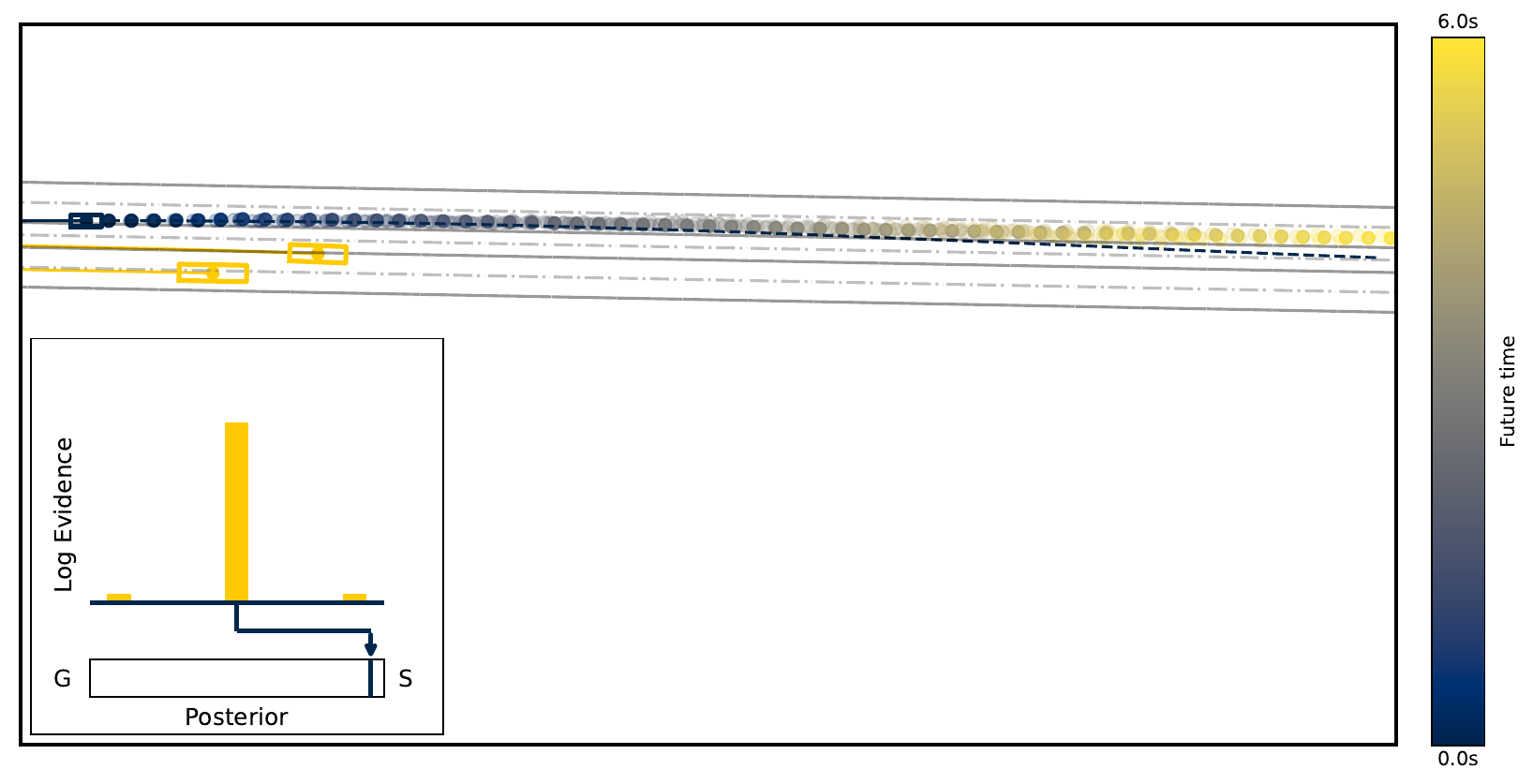}%
        \label{fig: visual f}} \\
        \subfloat[D3-P1]{\includegraphics[trim=0 0 60 0, clip, height=1.495in]{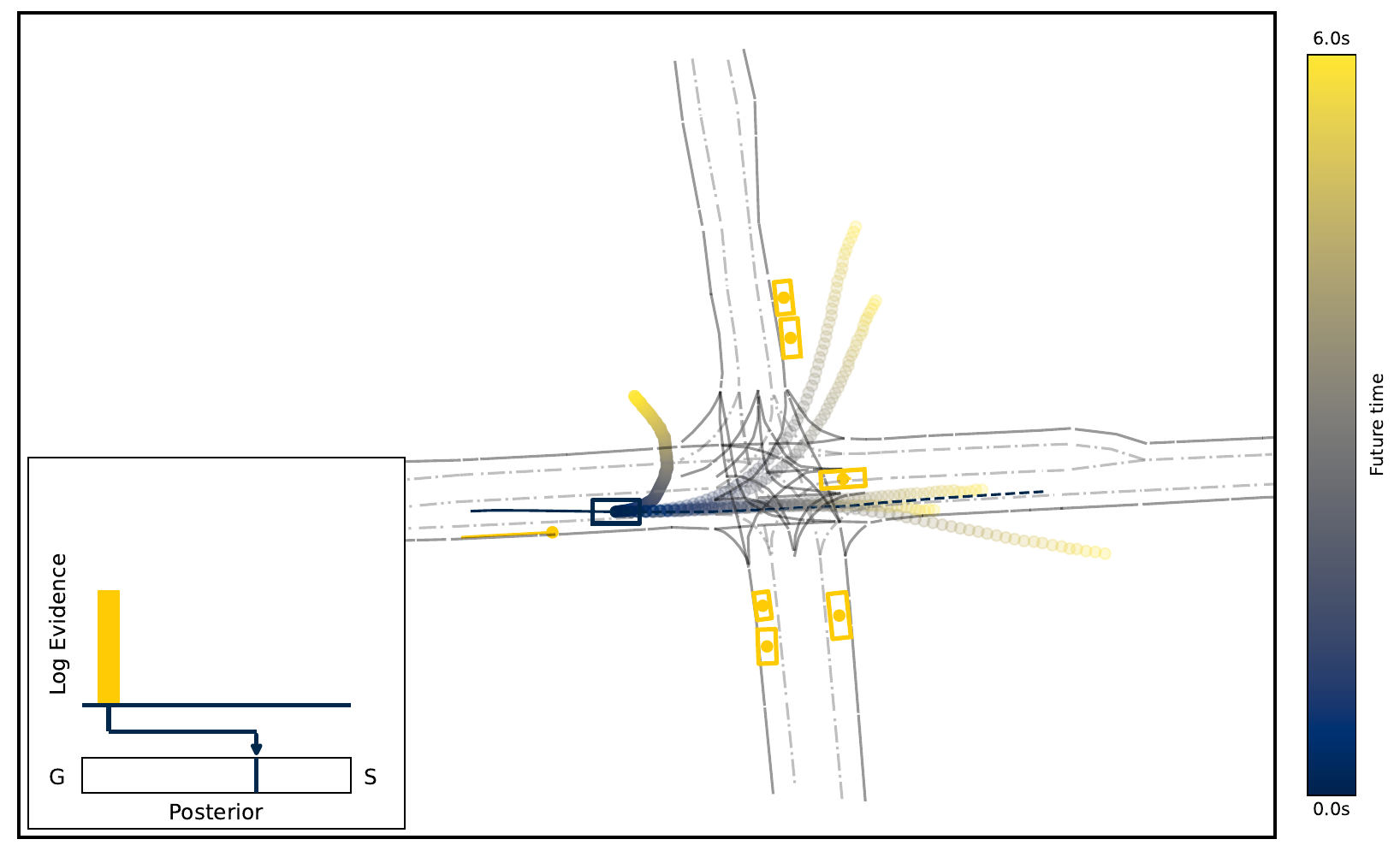}%
        \label{fig: visual g}}
        \subfloat[D3-P2]{\includegraphics[trim=0 0 60 0, clip, height=1.495in]{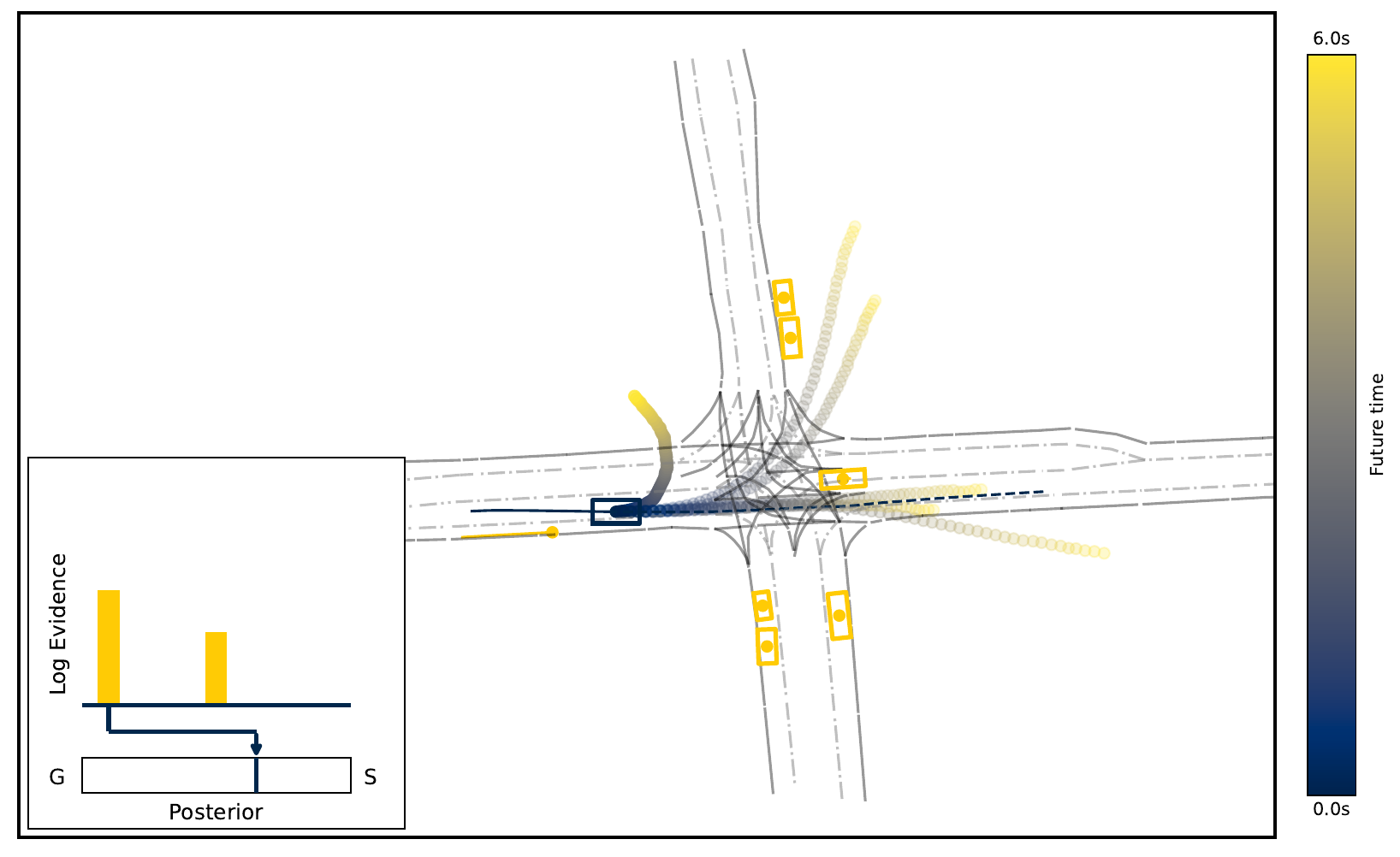}%
        \label{fig: visual h}}
        \subfloat[D3-P3]{\includegraphics[height=1.495in]{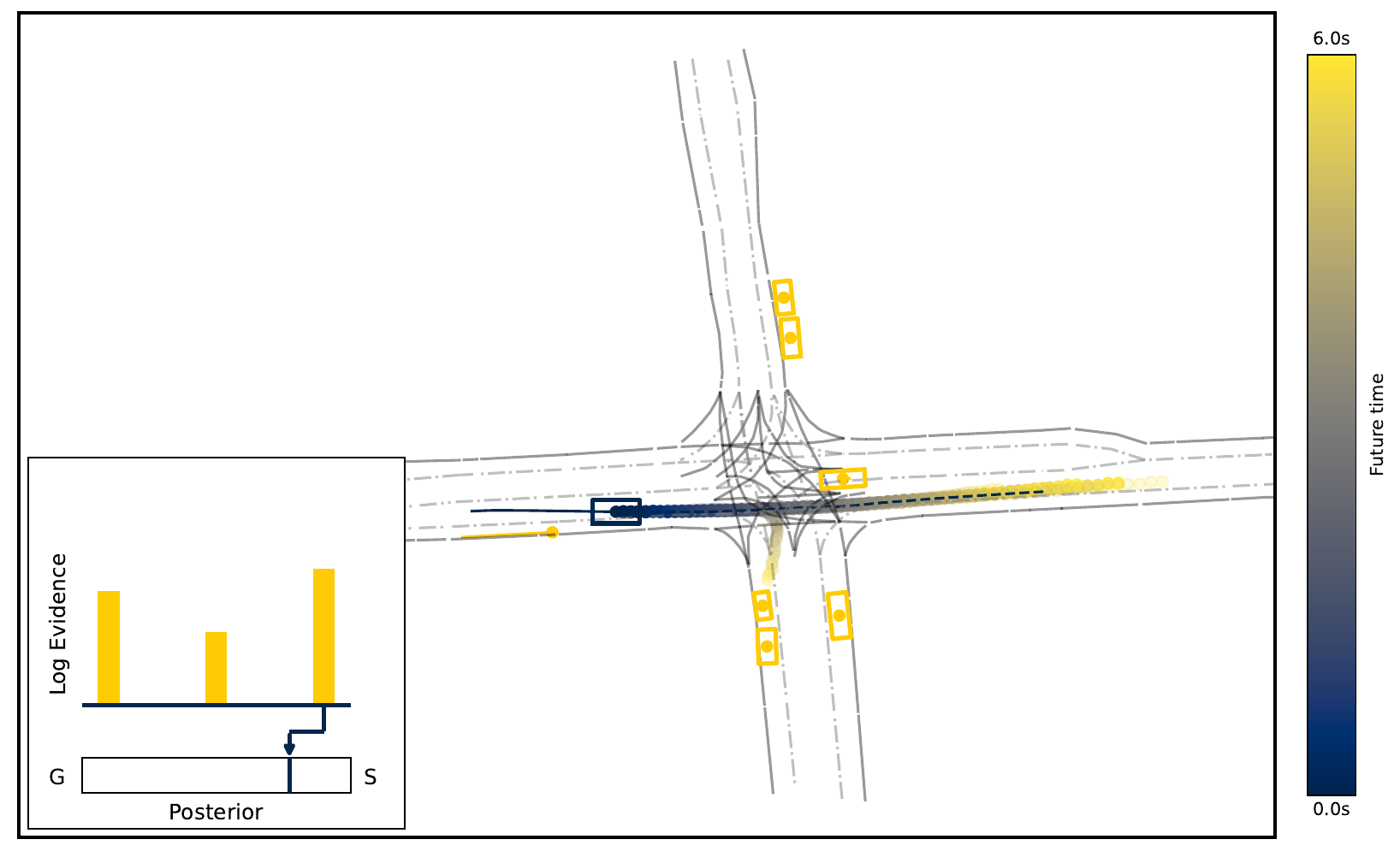}%
        \label{fig: visual i}} 
      \caption{Visualization of qualitative performance across different learning phases in the DECODE framework: Each row represents repeated observations of the same scene to highlight the consistency and variation in the model's predictive behavior as it transitions through phases. Columns delineate the different learning phases, demonstrating how predictions evolve with the progression of training and adaptation to new information. In each subfigure, yellow bars represent the log evidence computed by the normalizing flow for each specialized model, while the posterior weights assigned to the generalized (G) and specialized (S) models are shown below.}
      \label{fig: visual}
\end{figure*}
Fig. \ref{fig:main results} illustrates the evolution of the minADE metric throughout the entire continual learning process. The graph reveals that minADE consistently decreases below the performance levels of the corresponding pre-trained generalized model after each learning phase transition. The shifts in color mark the progression into new learning phases, effectively showcasing the model’s adaptability and performance improvement as it is sequentially exposed to different domains.

As shown in Table \ref{tab:table2}, our DECODE framework achieves performance comparable to or better than SOTA PET–based continual learning methods. Both $\texttt{DualPrompt}$ and $\texttt{CODA-P}$ were originally designed for image classification with a ViT-B/16 \cite{dosovitskiy2021an} backbone that uses only an encoder with self-attention layers, To ensure fair comparison, we adapted their prompting designs to fit the encoder–decoder MTR structure: for $\texttt{DualPrompt}$, we followed the original G-Prompt and E-Prompt configurations—placing G-Prompts in the first two encoder layers and E-Prompts in the last four—and added a variant extending E-Prompts to all six decoder layers; for $\texttt{CODA-P}$, we tested encoder-only, decoder-only, and encoder–decoder variants with prompt components inserted at corresponding stages. Both methods used the same prompt hyperparameters reported in their original papers. As summarized in Table \ref{tab:table3}, our DECODE framework consistently outperforms or matches them across all settings, demonstrating superior adaptability and effectiveness for continual learning in motion prediction.

\begin{table}[!b]
\caption{Performance comparison of PET variants\label{tab:table3}}
\centering
\setlength{\tabcolsep}{2.8pt}
\begin{adjustbox}{max width=\columnwidth}
\begin{tabular}{c|c|c|c}
\toprule[2pt]
\hline
\multicolumn{2}{c|}{Method} & $\tt{AER}\downarrow$ & $\tt{FGT}\downarrow$ \\
\multicolumn{2}{c|}{} & ($\tt{minADE}$ / $\tt{minFDE}$) & ($\tt{minADE}$ / $\tt{minFDE}$) \\
\hline
\multirow{2}{*}{$\texttt{Dualprompt}$} &$\texttt{enc}$ & $0.794\pm0.010$ / $2.053\pm0.029$ & $0.054\pm0.006$ / $0.027\pm0.012$  \\
&$\texttt{enc+dec}$ & $0.608\pm0.009$ / $1.509\pm0.037$ & $\mathbf{0.040\pm0.002}$ / $0.062\pm0.009$  \\
\hline
\multirow{2}{*}{$\texttt{Coda-P}$} & $\texttt{enc}$ & $0.721\pm0.007$ / $1.871\pm0.015$ & $0.255\pm0.008$ / $0.541\pm0.021$  \\
& $\texttt{dec}$ & $0.647\pm0.006$ / $1.633\pm0.018$ & $0.280\pm0.006$ / $0.698\pm0.017$  \\
&$\texttt{enc+dec}$ & $0.685\pm0.006$ / $1.728\pm0.017$ & $0.415\pm0.007$ / $1.020\pm0.016$  \\
\hline
\multicolumn{2}{c|}{\texttt{DECODE}} 
& $\mathbf{0.584\pm0.005} / \mathbf{1.395\pm0.019}$
& $0.044\pm0.001 / \mathbf{0.027\pm0.001}$ \\
\hline
\bottomrule[2pt]
\end{tabular}
\end{adjustbox}
\end{table}

Next, we present qualitative analysis by visualizing motion prediction results across three different learning phases for each of the three domains as in Fig \ref{fig: visual}. Future timesteps are depicted using a color gradient, with transparency levels indicating the probability of each modality. The log evidence, computed by a specialized normalizing flow, is shown on the left side of each visualization, accompanied by a box indicating the relative weighting of the posterior output—revealing whether the prediction aligns more with the pre-trained generalized model or the corresponding specialized model.

In Fig. \ref{fig: visual a} - \ref{fig: visual c} for the rounD domain, we observe that, once the specialized model is trained, the log evidence for the correct specialized model consistently dominates. The predicted trajectory remains consistent as the framework expands to the next domain, accurately covering the two major intentions: driving inside the circle and exiting at the subsequent roundabout exit with the six predicted trajectories.

In the HighD domain, as depicted in Fig. \ref{fig: visual d}, the DECODE framework handles scenes where no specialized model initially captures the scene correctly. The log evidence for the first specialized model is correctly estimated to be low, indicating unfamiliarity with the scene. Consequently, the output resembles that of the generalized model more closely. Despite some erroneous turning modalities, the generalized model's broad knowledge base allows for correct high-speed scenario predictions. However, by the end of learning phase 2, as shown in Fig. \ref{fig: visual e}, the correct specialized model is selected, resulting in significantly improved predictions. This learned knowledge is retained through the subsequent learning phase, as demonstrated in Fig. \ref{fig: visual f}.

Similarly, for the InD domain depicted in Fig. \ref{fig: visual g} and Fig. \ref{fig: visual h}, predictions made before the expansion show that the RounD specialized model has higher log evidence due to similarities between roundabouts and intersections. Upon completion of learning phase 3, the correct modalities—driving straight and turning right—are accurately captured as in Fig. \ref{fig: visual i}, further enhancing prediction quality.

\begin{figure}[!t]
\centering
\includegraphics[width=3.0in]{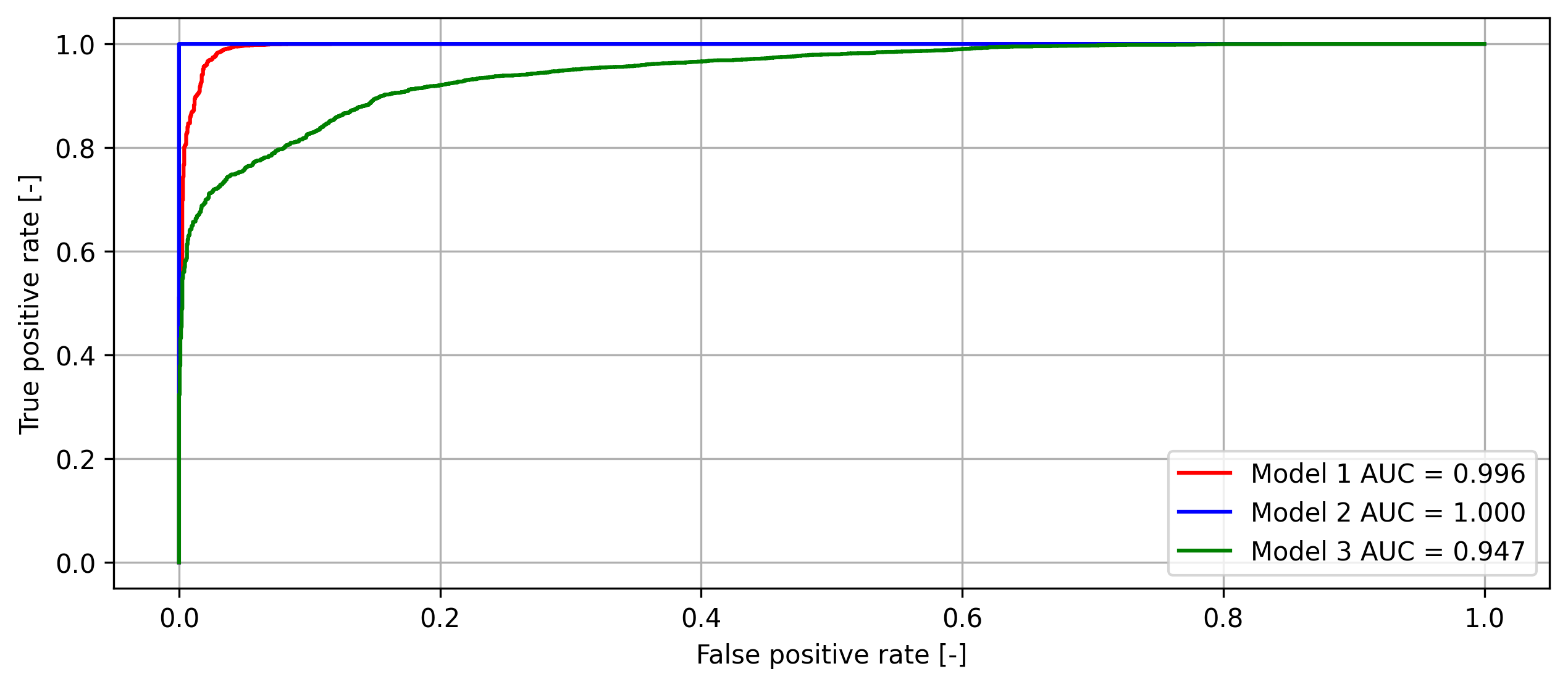}
\caption{ROC curves of three domain specialized normalizing flows}
\label{fig: roc}
\end{figure}
\begin{figure}[!t]
      \centering
        \subfloat[]{\includegraphics[trim=0 0 80 0, clip, height=1.45in]{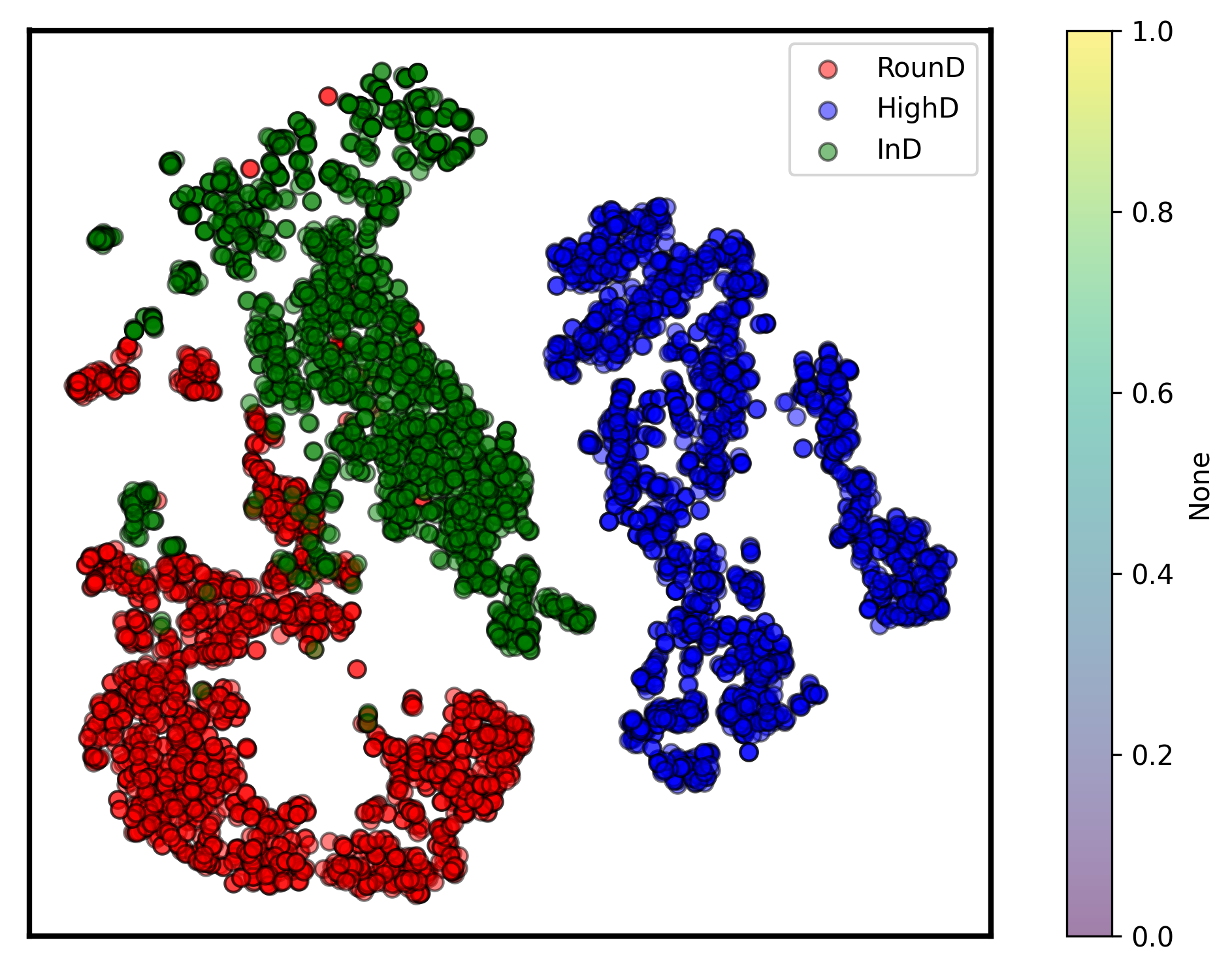}%
        \label{fig: tsne a}}    
        \subfloat[]{\includegraphics[height=1.45in]{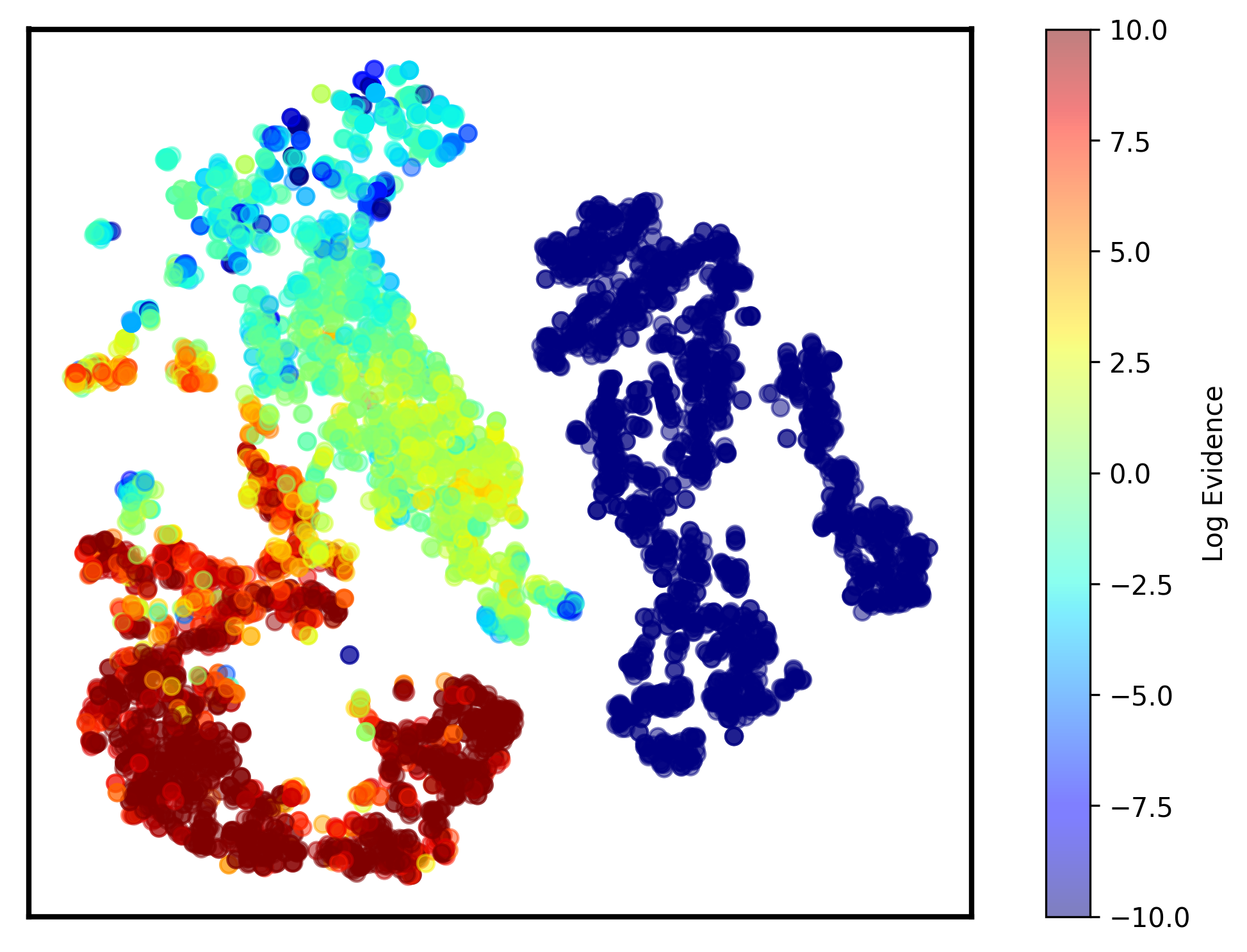}%
        \label{fig: tsne b}}\\
        \subfloat[]{\includegraphics[trim=0 0 80 0, clip, height=1.45in]{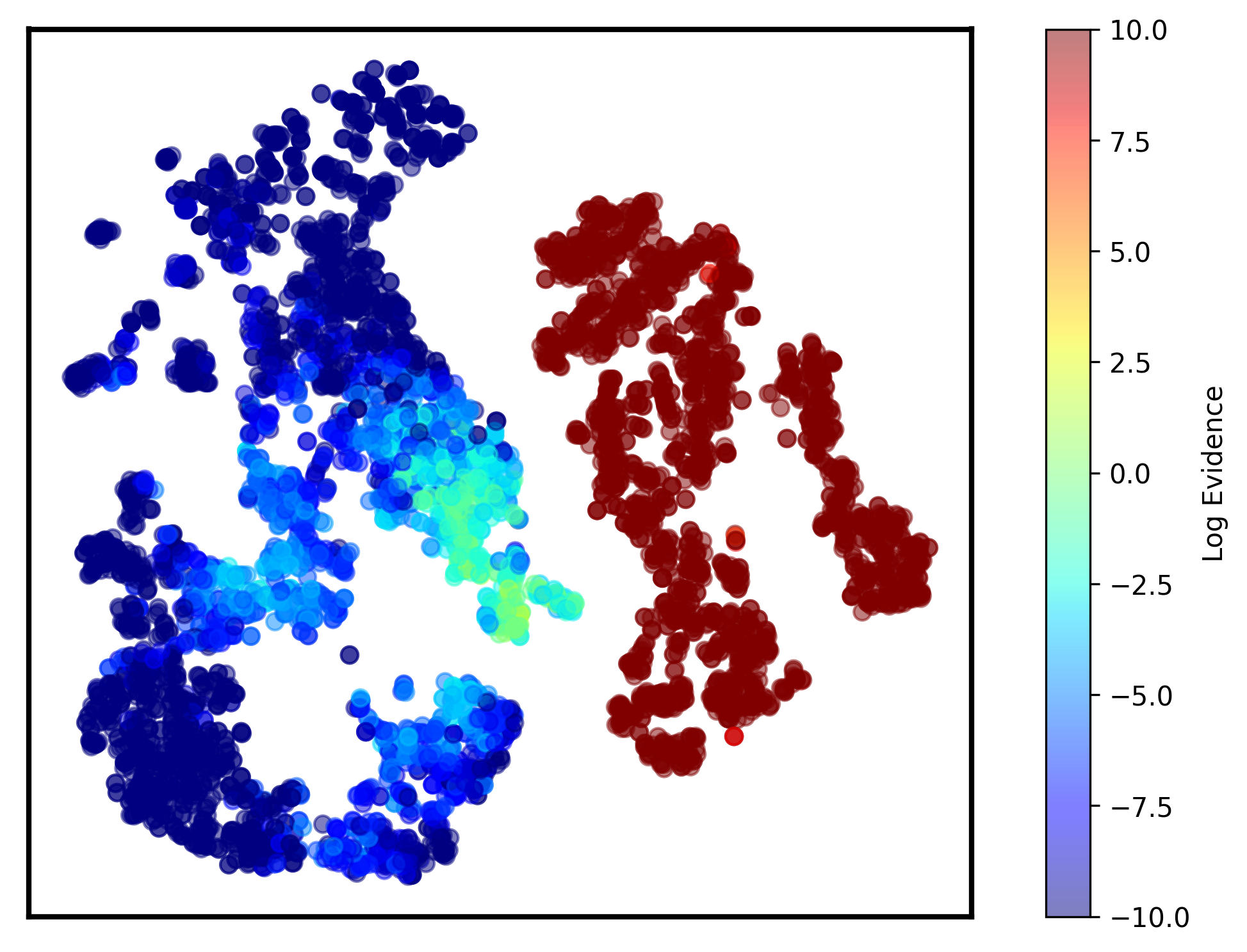}%
        \label{fig: tsne c}}
        \subfloat[]{\includegraphics[height=1.45in]{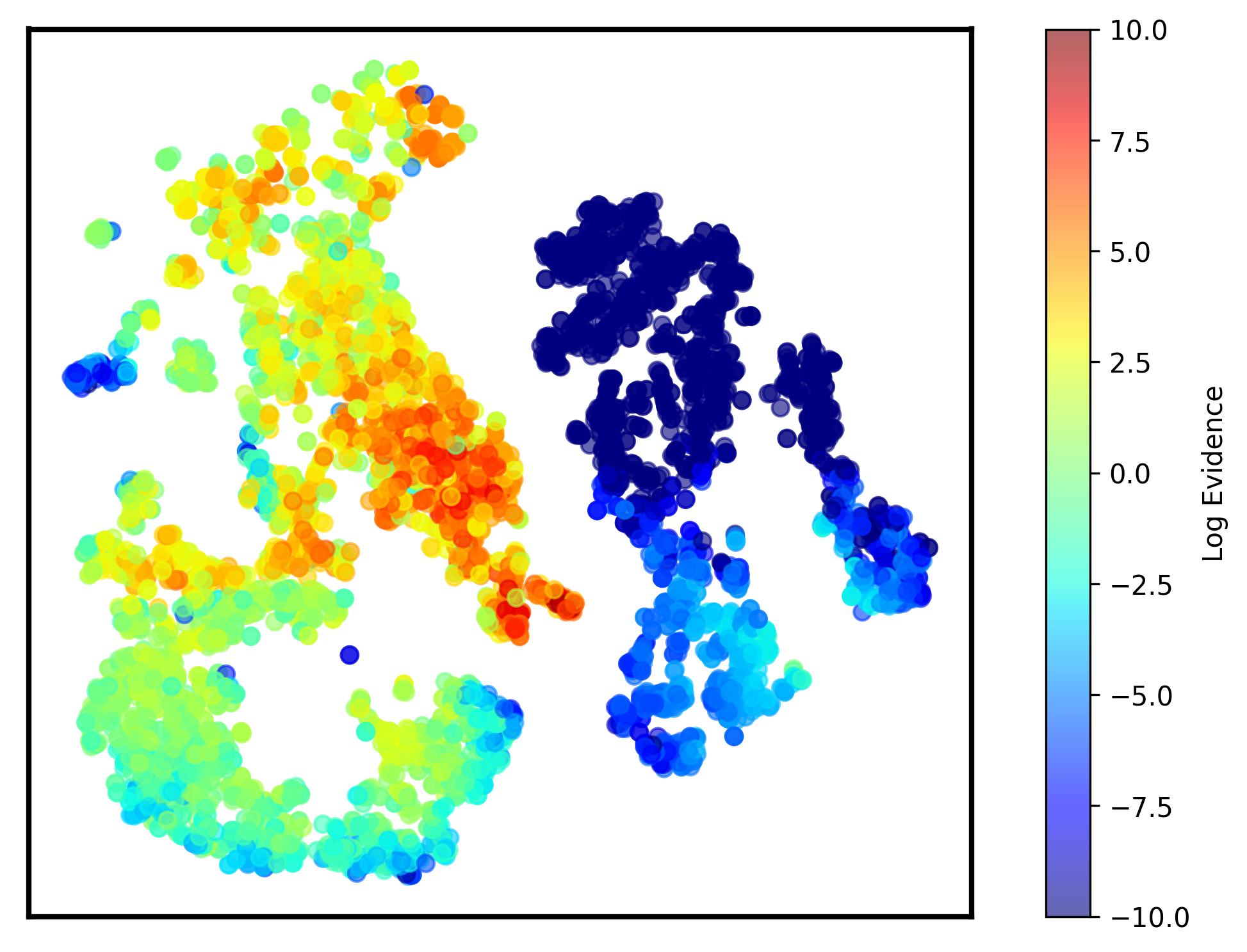}%
        \label{fig: tsne d}}
        
      \caption{Visualization of hidden representations using t-SNE: (a) illustrates the ground truth labels of the domains of the scenes, with different colors representing each domain: red for RounD ($\tt{D1}$), blue for HighD ($\tt{D2}$), and green for InD ($\tt{D3}$). (b)-(d) log evidence $\log\mathbb{P}_h$ computed respectively by the $\tt{D1}$, $\tt{D2}$, and $\tt{D3}$ specialized normalizing flow}
      \label{fig: tsne log evidence}
\end{figure}
Next, we evaluate the domain awareness capabilities of our specialized normalizing flow models. Given that the evidence estimated by these models is crucial in selecting the most appropriate specialized model for a given scene, we first approach this as a binary classification problem. Here, the negative log-likelihood output by each normalizing flow model classifies whether a scene is outside its specialized domain (labeled as 1) or within it (labeled as 0). To ensure a balanced comparison, approximately 2000 validation scenarios from each domain are randomly selected.

Each normalizing flow model is finely tuned to identify scenarios that do not belong to its learned domain. Fig. \ref{fig: roc} illustrates the Receiver Operating Characteristic (ROC) curve for the three domain-specialized normalizing flow models after the third learning phase, highlighting the trade-off between the true positive rate and the false positive rate. Ideally, a more accurate model will depict a curve closer to the top left corner of the graph. The Area Under the ROC Curve (AUROC) is also calculated for each model, with a score of 1.0 indicating perfect classification and 0.5 or less indicating no discriminatory ability. As evidenced in the figure, all three normalizing flow models demonstrate robust classification capabilities, achieving an average AUROC of 0.988.
 
For qualitative analysis, we examine the universal hidden representations $h$ generated by the pre-trained generalized MTR encoder using WOMD data. Given that these hidden features are high-dimensional (256 dimensions), direct visualization is impractical. Therefore, we employ the t-SNE technique to project these features into a two-dimensional space, as illustrated in Fig. \ref{fig: tsne log evidence}. Approximately 2000 scenes from each domain are randomly selected, and the resulting t-SNE plots clearly show clusters corresponding to different domains, as shown in Fig. \ref{fig: tsne a}. This clustering underscores the generalizability of the universal hidden representations.

Further, we assess the estimated log evidence $\log\mathbb{P}_h$ produced by each domain-specialized normalizing flow across all scenes, applying a color map to indicate estimated familiarity among the scenes. For the RounD domain model as in Fig. \ref{fig: tsne b}, a clear distinction is evident between roundabout scenarios (red) and highway scenarios (blue), with some intersection scenes exhibiting similarities to roundabouts. In Fig. \ref{fig: tsne c}, the HighD domain model shows a sharper distinction, clearly identifying representations for highways, with only a small fraction of intersection scenarios closer to highways. The InD domain model, however, exhibits less discriminative ability compared to the last two; a notable portion of the intersection scenarios display a color similar to that of the roundabout scenarios as shown in Fig. \ref{fig: tsne d}.
\begin{table}[!b]
    \centering
    \caption{Confusion Matrix of Three Learning Phases}
    \begin{tabular}{|c|c|c|c|c|}
        \toprule[2pt]
        \hline
        \multicolumn{2}{|c|}{} & \multicolumn{3}{c|}{Predicted} \\ \cline{3-5}
        \multicolumn{2}{|c|}{} & RounD & HighD & InD \\ \hline
        \multirow{3}{*}{\rotatebox{90}{Actual}} & RounD & 2011 & 0 & 2 \\ \cline{2-5}
        & HighD & 0 & 2016 & 0 \\ \cline{2-5}
        & InD & 10 & 0 & 1962 \\ \hline
        \bottomrule[2pt]
    \end{tabular}
    \label{table: confusion 3}
\end{table}
\begin{table}[!b]
    \centering
    \caption{Confusion Matrix of Normal vs. Conflict Domains}
    \begin{tabular}{|c|c|c|c|}
        \toprule[2pt]
        \hline
        \multicolumn{2}{|c|}{} & \multicolumn{2}{c|}{Predicted} \\ \cline{3-4}
        \multicolumn{2}{|c|}{} & Normal & Conflict \\ \hline
        \multirow{2}{*}{Actual} & Normal & 1951 & 25 \\ \cline{2-4}
        & Conflict & 9 & 504 \\ \hline
        \bottomrule[2pt]
    \end{tabular}
    \label{table: confusion 2}
\end{table}
Following the procedure outlined in Algorithm 1, the most suitable specialized model is selected based on the highest log evidence generated by the normalizing flow models. Table \ref{table: confusion 3} presents a confusion matrix for the RounD, HighD, and InD domains, illustrating the classification performance across different validation sets. The data in the table indicate that the majority of scenes are correctly classified, with the predictions being generated by the specialized model that is most familiar with each respective scene. The overall classification accuracy is exceptionally high, with an accuracy of 0.998, a precision of 0.998, and a recall of 0.997, demonstrating the effectiveness of our model selection process based on domain familiarity.

We have further extended our model to encompass domains not strictly defined by geographical locations. To validate the efficacy of our approach, we conducted simulations within the Mcity test environment, a high-fidelity urban simulation featuring a 1,000-foot arterial, six signalized intersections, three roundabouts, and a mix of single and multi-lane roads. We employed a Naturalistic Driving Environment to generate realistic driving scenarios that include both normal driving conditions and conflict events, thereby creating a normal domain dataset and a conflict domain dataset. The latter involves the same locations as the normal domain but exhibits varying levels of vehicle aggressiveness.

To accurately model human driving behavior under typical conditions, we calibrated car-following and lane-changing models using Naturalistic Driving Data. Additionally, we incorporated mechanisms to simulate human negligence and to replicate diverse safety-critical scenarios, aligning with real-world occurrence rates and patterns \cite{feng2021intelligent}.

Initially, training commenced with the normal domain and subsequently expanded to include the conflict domain. Table \ref{table: confusion 2} presents the confusion matrix for the normalizing flow models, demonstrating an accuracy of 0.986, a precision of 0.974, and a recall of 0.985. These results highlight the model’s capability to effectively distinguish between different vehicular behaviors, which is essential for providing crucial safety information in the planning task.

\begin{table}[!b]
\caption{Comparison between models with and without deep Bayesian uncertainty estimation\label{table: prior e}}
\centering
\begin{tabular}{c|c|c|c}
\toprule[2pt]
\hline
verion & RounD & HighD & InD \\
&($\tt{minADE}$/$\tt{minFDE}$)&($\tt{minADE}$/$\tt{minFDE}$)&($\tt{minADE}$/$\tt{minFDE}$) \\
\hline
w/o & $0.527/1.272$ & $65.87/130.88$ & $1.984/5.514$ \\
$e^{(0)}=10$ & $0.531/1.270$ & $0.839/2.238$ & $1.165/3.089$ \\
\hline
\bottomrule[2pt]
\end{tabular}
\end{table}

\subsection{Ablation Study}
\begin{figure}[!t]
      \centering
        \subfloat[]{\includegraphics[height=1.42in]{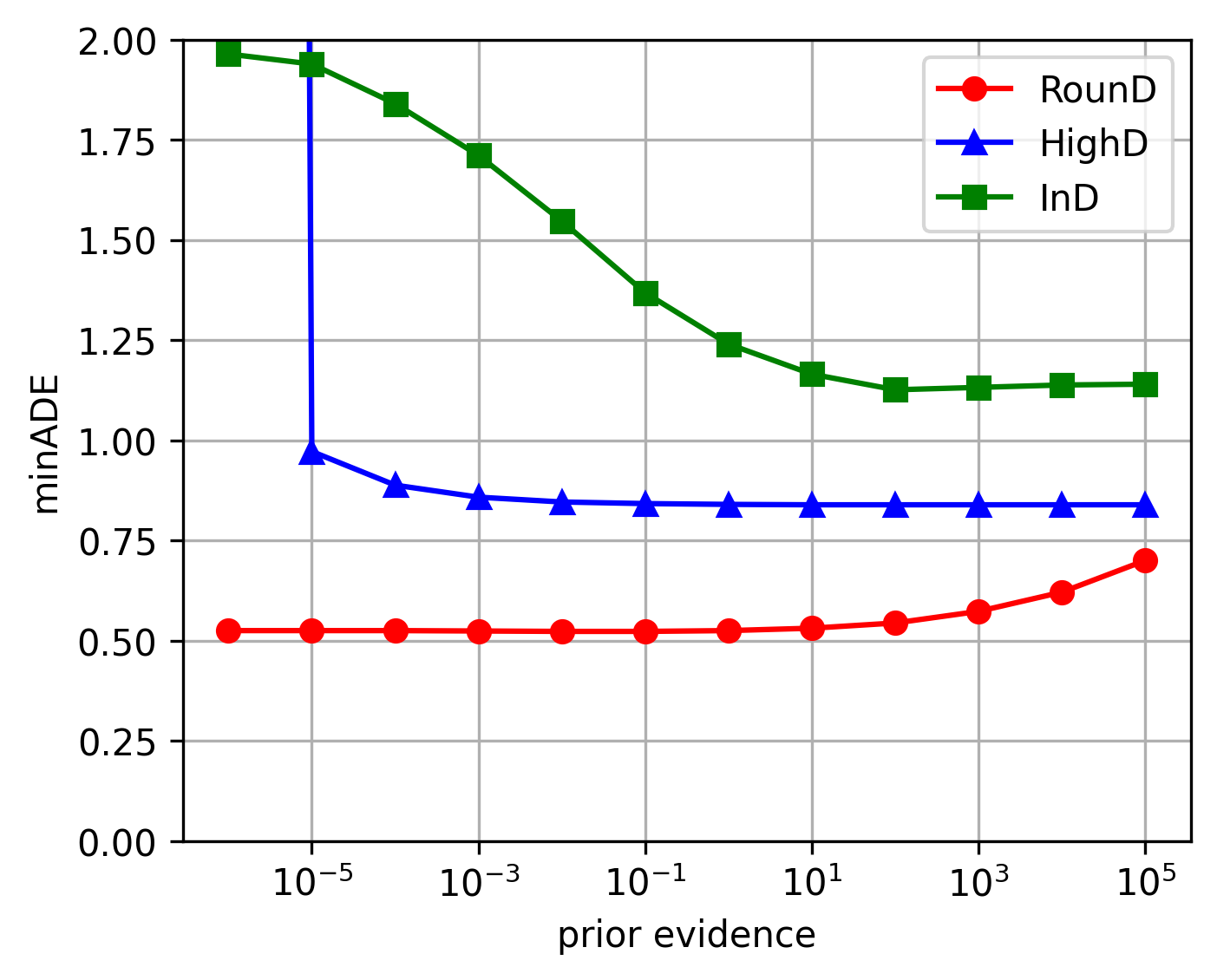}%
        \label{fig: prior ade}}    
        \hfil
        \subfloat[]{\includegraphics[height=1.42in]{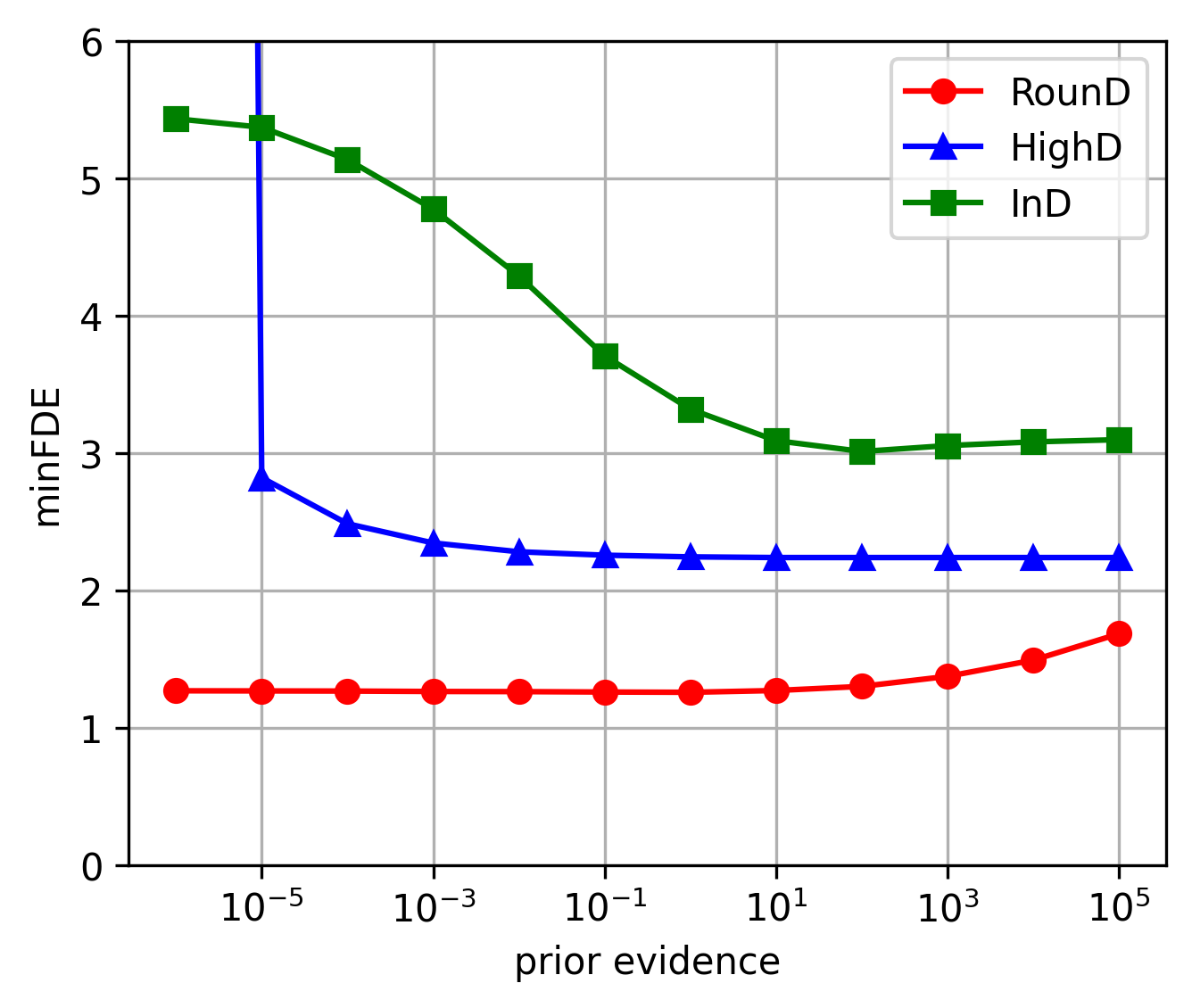}%
        \label{fig: prior fde}}
        
      \caption{Sensitivity of different choices of prior evidence $e^{(0)}$ on the performance of different domains}
      \label{fig: prior evidence}
\end{figure}

To evaluate the impact of dynamic balancing between generalized and specialized models using deep Bayesian uncertainty estimation, Table \ref{table: prior e} presents a comparison of our proposed model's performance with and without this feature. We analyze the model after phase 1, which includes only one specialized model trained on the RounD domain, assessing its performance on the HighD and InD domains. For the dynamic balancing variant, we set the prior evidence at $10$. The results indicate improved performance in both domains when dynamic balancing is implemented, with particularly notable enhancements observed in the HighD domain. This improvement substantiates the necessity of dynamic balancing for the framework to manage unfamiliar scenarios effectively. Further analysis of the sensitivity of various prior evidence settings is depicted in Fig. \ref{fig: prior evidence}. Increasing prior evidence generally enhances performance in both the HighD and InD domains until it stabilizes. The HighD domain exhibits a significant initial decrease in minADE and minFDE. However, as prior evidence continues to increase, the performance in the RounD domain deteriorates, indicating an over-reliance on the generalized model even though the specialized model shows superior results. In practice, we follow a simple hyperparameter tuning strategy to choose the value of the prior evidence. Specifically, we incrementally increase $e^{(0)}$ starting from zero and select the value that yields the best performance on the main evaluation metric (minADE) over a small buffer of representative samples from each domain, assuming access to limited validation data. This approach provides a practical and effective guideline for setting the prior evidence in a data-driven manner. 

\subsection{Computational and Parameter Efficiency Analysis}
\begin{figure}[!t]
      \centering
        \subfloat[]{\includegraphics[height=1.7in]{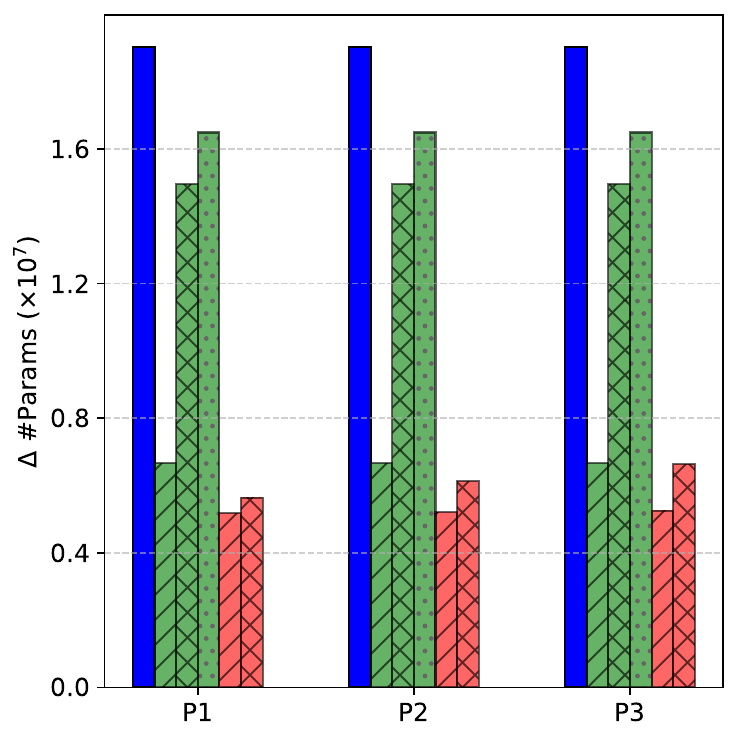}%
        \label{fig: param_growth}}    
        \hfil
        \subfloat[]{\includegraphics[height=1.7in]{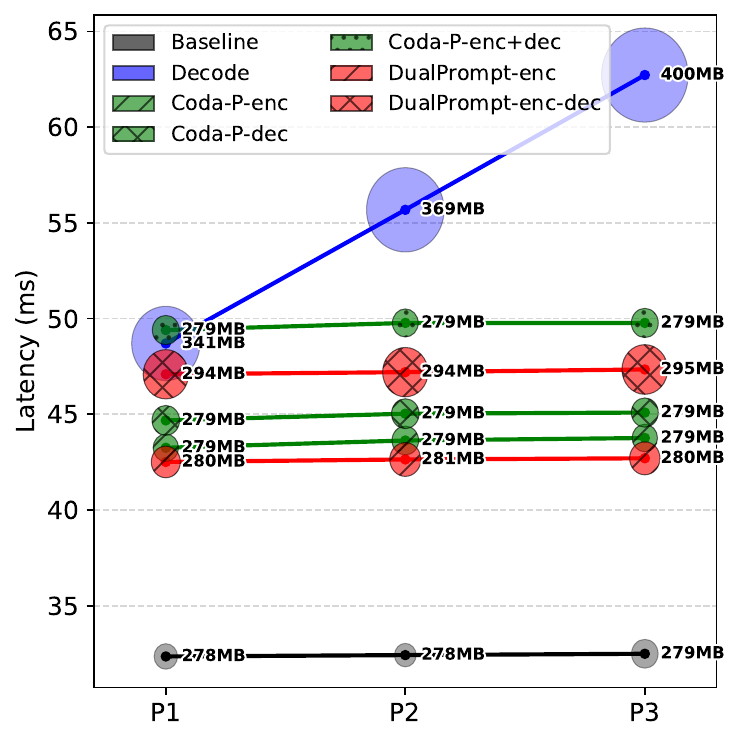}%
        \label{fig: latency}}
        
      \caption{Parameter and runtime efficiency comparison among continual learning methods across different learning phases. 
(a) Parameter growth of each method, measured as the overhead relative to the baseline model (MTR). 
(b) Inference latency (line plot) and GPU memory footprint (circle size), illustrating the runtime overhead of each method.}
      \label{fig: overhead}
\end{figure}

We further compare the parameter and runtime efficiency. As shown in Fig.~\ref{fig: param_growth}, the parameter overhead and growth pattern across learning phases are illustrated. We compare DECODE with all $\texttt{DualPrompt}$ and $\texttt{CODA-P}$ variants. The rest methods maintain the same parameter count as the original MTR model. Our framework introduces the largest initial parameter overhead due to the inclusion of a hypernetwork, but its growth across learning phases is negligible since only a small domain query component is added per new task. In contrast, although $\texttt{DualPrompt}$ begins with minimal overhead, it requires progressively larger prompt sets as new tasks are introduced. $\texttt{CODA-P}$ maintains a fixed prompt pool and exhibits a parameter count comparable to DECODE.

Fig.~\ref{fig: latency} presents the inference latency per frame and GPU memory usage across learning phases. For a fair comparison, we evaluate on the rounD domain with up to 8 target vehicles per frame. While DECODE initially exhibits similar latency to other methods, its computation time grows linearly and faster as the number of domains increases. This is primarily due to the need to evaluate the normalizing flow for each known specialized domain during inference.

To address this limitation, we highlight two future optimization directions:
\begin{itemize}
    \item Hypernetwork optimization — our current implementation performs model-wise looping instead of batching across domains, which could be significantly accelerated.
    \item Domain pruning — since domain shifts between consecutive frames are typically small, the set of candidate domains can be constrained based on recent similarity metrics. Additionally, organizing domains into a hierarchical structure could make best-domain selection more scalable as the number of specialized models grows.
\end{itemize}

\subsection{Extension to General Continual Learning Tasks}
\begin{table}[b!]
\caption{Results on 5-datasets (10 classes per task)\label{tab:table6}}
\centering
\begin{threeparttable}
\begin{tabular}{c|c|c}
\toprule[2pt]
\hline
Method & $\tt{Avg. ACC}\uparrow$ & $\tt{FGT}\downarrow$ \\
\hline
$\texttt{ER(500)}$ & $84.26\pm0.84^\dagger$ & $12.85\pm0.62^\dagger$  \\
$\texttt{DER++(500)}$ & $84.88\pm0.57^\dagger$ & $10.46\pm1.02^\dagger$  \\
$\texttt{EWC}$ & $50.93\pm0.09^\dagger$ & $34.94\pm0.07^\dagger$  \\
$\texttt{LwF}$ & $47.91\pm0.33^\dagger$ & $38.01\pm0.28^\dagger$  \\
$\texttt{F2P}$ & $81.14\pm0.93^\dagger$ & $4.64\pm0.52^\dagger$  \\
$\texttt{Dualprompt}$ & $88.08\pm0.36^\dagger$ & $2.21\pm0.69^\dagger$  \\
$\texttt{Coda-P}$ & $84.82\pm0.45^\ddagger$ & $13.24\pm0.45^\ddagger$  \\
$\texttt{DECODE}$ & $\mathbf{90.56\pm1.43}$ & $\mathbf{0.05\pm0.01}$  \\
\hline
\bottomrule[2pt]
\end{tabular}
\begin{tablenotes}
\footnotesize
\item[$\dagger$Reported by \cite{wang2022dualprompt}. $\ddagger$ Reproduced using original codebase \cite{smith2023coda}.] 
\end{tablenotes}
\end{threeparttable}
\end{table}

\begin{table}[!b]
\caption{Results on 10-task CIFAR-100 (10 classes per task)\label{tab:table7}}
\centering
\begin{threeparttable}
\begin{tabular}{c|c|c}
\toprule[2pt]
\hline
Method & $\tt{Avg. ACC}\uparrow$ & $\tt{FGT}\downarrow$ \\
\hline
$\texttt{ER(5000)}$ & $82.53\pm0.17^\dagger$ & $16.46\pm0.25^\dagger$  \\
$\texttt{DER++(5000)}$ & $83.94\pm0.34^\dagger$ & $14.55\pm0.73^\dagger$  \\
$\texttt{EWC}$ & $47.01\pm0.29^\dagger$ & $33.27\pm1.17^\dagger$  \\
$\texttt{LwF}$ & $60.69\pm0.63^\dagger$ & $27.77\pm2.17^\dagger$  \\
$\texttt{F2P}$ & $83.86\pm0.28^\dagger$ & $7.35\pm0.38^\dagger$  \\
$\texttt{Dualprompt}$ & $86.51\pm0.33^\dagger$ & $5.16\pm0.09^\dagger$  \\
$\texttt{Coda-P}$ & $\mathbf{91.71\pm0.78}^\ddagger$& $\mathbf{3.26\pm0.83}^\ddagger$  \\
$\texttt{DECODE}$ & $86.41\pm1.23$ & $4.55\pm0.51$  \\
\hline
\bottomrule[2pt]
\end{tabular}
\begin{tablenotes}
\footnotesize
\item[$\dagger$Reported by \cite{wang2022dualprompt}. $\ddagger$ Reproduced using original codebase \cite{smith2023coda}.]
\end{tablenotes}
\end{threeparttable}
\end{table}

The DECODE framework is not limited to motion prediction and can be extended to general continual learning applications, such as image classification, which is widely studied in the literature. To demonstrate this, we adapt DECODE to work with a ViT-B/16 model \cite{dosovitskiy2021an} pre-trained on the ImageNet-1k \cite{ILSVRC15} dataset, using it as the backbone and generalized model. Unlike motion prediction models that often adopt an encoder–decoder architecture, the ViT-B/16 model consists of an encoder followed by a classification head. To integrate DECODE, we make the following modifications:

We consider the encoder with $K = 12$ self-attention layers, and define two parallel computational routes. The first route passes the final ($K$-th) layer’s output to produce a [CLS] token, which is then used as input to the normalizing flow for domain inference. The second route branches from the $(K{-}1)$-th layer and connects to a specialized self-attention layer followed by a specialized classification head, whose parameters are generated by the hypernetwork. This design preserves the core structure of the original ViT model while enabling domain specialization, demonstrating the flexibility of our framework to support task-specific adaptation across diverse continual learning scenarios. 

Table~\ref{tab:table6} presents results on a five-dataset benchmark \cite{ebrahimi2020adversarial} composed of diverse image classification tasks: CIFAR-10\cite{cifar10}, MNIST\cite{mnist}, SVHN\cite{svhn}, notMNIST\cite{notmnist}, and Fashion-MNIST\cite{fashion}. This setting reflects a realistic continual learning scenario with high task heterogeneity. Table~\ref{tab:table7} reports results on the widely used 10-task CIFAR-100 benchmark \cite{cifar10}, a standard for evaluating class-incremental learning. During training, we freeze the pretrained ViT backbone in the same way as the prompt-based baselines used for comparison, and only train the hypernetwork. All methods are trained with the same data exposure and number of optimization steps (25 epochs for five-dataset and 40 epochs for CIFAR-100), ensuring comparable training budgets. For each experiment, we conduct five training trials and report the average accuracy and forgetting, following the definitions in \cite{wang2022dualprompt}. Across both benchmarks, our DECODE framework consistently achieves competitive or superior performance compared to state-of-the-art continual learning methods. Notably, DECODE performs particularly well on the five-dataset benchmark, where we observe that the normalizing flow is more effective when task diversity is high.

\section{Conclusion and Future Work}
In conclusion, the DECODE framework introduces a robust and scalable solution to the challenges of continual learning in motion prediction for autonomous vehicles. By integrating hypernetworks and normalizing flow models, DECODE efficiently manages the complexity of training on continually evolving data without substantial parameter expansion. Our experimental results validate the framework's ability to maintain a low forgetting rate and achieve impressive accuracy across different domains, significantly outperforming traditional active learning strategies. The utilization of deep Bayesian uncertainty estimation further enhances the reliability of predictions, ensuring that outputs are both robust and aligned with the specific demands of varied and unpredictable driving environments.

Looking ahead, we plan to enhance DECODE by optimizing the hypernetwork architecture—keeping parameters fixed while expanding through new domain queries. We also aim to improve the efficiency of the domain awareness module by batching operations and bounding the number of active domains per frame to reduce latency. These efforts will further broaden DECODE’s adaptability and precision.

\bibliographystyle{IEEEtran}
\bibliography{ref}

@article{cheng2024pluto,
  title={PLUTO: Pushing the Limit of Imitation Learning-based Planning for Autonomous Driving},
  author={Cheng, Jie and Chen, Yingbing and Chen, Qifeng},
  journal={arXiv preprint arXiv:2404.14327},
  year={2024},
}

@article{fridovich2020confidence,
  title={Confidence-aware motion prediction for real-time collision avoidance1},
  author={Fridovich-Keil, David and Bajcsy, Andrea and Fisac, Jaime F and Herbert, Sylvia L and Wang, Steven and Dragan, Anca D and Tomlin, Claire J},
  journal={The International Journal of Robotics Research},
  volume={39},
  number={2-3},
  pages={250--265},
  year={2020},
  publisher={SAGE Publications Sage UK: London, England}
}

@article{foka2010probabilistic,
  title={Probabilistic autonomous robot navigation in dynamic environments with human motion prediction},
  author={Foka, Amalia F and Trahanias, Panos E},
  journal={International Journal of Social Robotics},
  volume={2},
  pages={79--94},
  year={2010},
  publisher={Springer}
}

@INPROCEEDINGS{9561666,
  author={Bergamini, Luca and Ye, Yawei and Scheel, Oliver and Chen, Long and Hu, Chih and Del Pero, Luca and Osiński, Błażej and Grimmett, Hugo and Ondruska, Peter},
  booktitle={2021 IEEE International Conference on Robotics and Automation (ICRA)}, 
  title={SimNet: Learning Reactive Self-driving Simulations from Real-world Observations}, 
  year={2021},
  volume={},
  number={},
  pages={5119-5125},
}

@article{yan2023learning,
  title={Learning naturalistic driving environment with statistical realism},
  author={Yan, Xintao and Zou, Zhengxia and Feng, Shuo and Zhu, Haojie and Sun, Haowei and Liu, Henry X},
  journal={Nature communications},
  volume={14},
  number={1},
  pages={2037},
  year={2023},
  publisher={Nature Publishing Group UK London}
}

@InProceedings{Chang_2019_CVPR,
author = {Chang, Ming-Fang and Lambert, John and Sangkloy, Patsorn and Singh, Jagjeet and Bak, Slawomir and Hartnett, Andrew and Wang, De and Carr, Peter and Lucey, Simon and Ramanan, Deva and Hays, James},
title = {Argoverse: 3D Tracking and Forecasting With Rich Maps},
booktitle = {Proceedings of the IEEE/CVF Conference on Computer Vision and Pattern Recognition (CVPR)},
month = {June},
year = {2019}
}

@article{zhan2019interaction,
  title={Interaction dataset: An international, adversarial and cooperative motion dataset in interactive driving scenarios with semantic maps},
  author={Zhan, Wei and others},
  journal={arXiv preprint arXiv:1910.03088},
  year={2019}
}

@inproceedings{ettinger2021large,
  title={Large scale interactive motion forecasting for autonomous driving: The waymo open motion dataset},
  author={Ettinger, Scott and Cheng, Shuyang and Caine, Benjamin and Liu, Chenxi and Zhao, Hang and Pradhan, Sabeek and Chai, Yuning and Sapp, Ben and Qi, Charles R and Zhou, Yin and others},
  booktitle={Proceedings of the IEEE/CVF International Conference on Computer Vision},
  pages={9710--9719},
  year={2021}
}

@article{feng2024unitraj,
  title={UniTraj: A Unified Framework for Scalable Vehicle Trajectory Prediction},
  author={Feng, Lan and Bahari, Mohammadhossein and Amor, Kaouther Messaoud Ben and Zablocki, {\'E}loi and Cord, Matthieu and Alahi, Alexandre},
  journal={arXiv preprint arXiv:2403.15098},
  year={2024}
}

@article{lefevre2014survey,
  title={A survey on motion prediction and risk assessment for intelligent vehicles},
  author={Lef{\`e}vre, St{\'e}phanie and Vasquez, Dizan and Laugier, Christian},
  journal={ROBOMECH journal},
  volume={1},
  pages={1--14},
  year={2014},
  publisher={Springer}
}

@InProceedings{Alahi_2016_CVPR,
author = {Alahi, Alexandre and Goel, Kratarth and Ramanathan, Vignesh and Robicquet, Alexandre and Fei-Fei, Li and Savarese, Silvio},
title = {Social LSTM: Human Trajectory Prediction in Crowded Spaces},
booktitle = {Proceedings of the IEEE Conference on Computer Vision and Pattern Recognition (CVPR)},
month = {June},
year = {2016}
}

@InProceedings{Gupta_2018_CVPR,
author = {Gupta, Agrim and Johnson, Justin and Fei-Fei, Li and Savarese, Silvio and Alahi, Alexandre},
title = {Social GAN: Socially Acceptable Trajectories With Generative Adversarial Networks},
booktitle = {Proceedings of the IEEE Conference on Computer Vision and Pattern Recognition (CVPR)},
month = {June},
year = {2018}
}

@article{chai2019multipath,
  title={Multipath: Multiple probabilistic anchor trajectory hypotheses for behavior prediction},
  author={Chai, Yuning and Sapp, Benjamin and Bansal, Mayank and Anguelov, Dragomir},
  journal={arXiv preprint arXiv:1910.05449},
  year={2019}
}

@article{tang2019multiple,
  title={Multiple futures prediction},
  author={Tang, Charlie and Salakhutdinov, Russ R},
  journal={Advances in neural information processing systems},
  volume={32},
  year={2019}
}

@inproceedings{salzmann2020trajectron++,
  title={Trajectron++: Dynamically-feasible trajectory forecasting with heterogeneous data},
  author={Salzmann, Tim and Ivanovic, Boris and Chakravarty, Punarjay and Pavone, Marco},
  booktitle={Computer Vision--ECCV 2020: 16th European Conference, Glasgow, UK, Proceedings, Part XVIII 16},
  pages={683--700},
  year={2020},
  organization={Springer}
}

@inproceedings{gao2020vectornet,
  title={Vectornet: Encoding hd maps and agent dynamics from vectorized representation},
  author={Gao, Jiyang and Sun, Chen and Zhao, Hang and Shen, Yi and Anguelov, Dragomir and Li, Congcong and Schmid, Cordelia},
  booktitle={Proceedings of the IEEE/CVF conference on computer vision and pattern recognition},
  pages={11525--11533},
  year={2020}
}

@inproceedings{liang2020learning,
  title={Learning lane graph representations for motion forecasting},
  author={Liang, Ming and Yang, Bin and Hu, Rui and Chen, Yun and Liao, Renjie and Feng, Song and Urtasun, Raquel},
  booktitle={Computer Vision--ECCV 2020: 16th European Conference, Glasgow, UK, Proceedings, Part II 16},
  year={2020},
  organization={Springer}
}

@INPROCEEDINGS{9812107,
  author={Varadarajan, Balakrishnan and Hefny, Ahmed and Srivastava, Avikalp and Refaat, Khaled S. and Nayakanti, Nigamaa and Cornman, Andre and Chen, Kan and Douillard, Bertrand and Lam, Chi Pang and Anguelov, Dragomir and Sapp, Benjamin},
  booktitle={2022 International Conference on Robotics and Automation (ICRA)}, 
  title={MultiPath++: Efficient Information Fusion and Trajectory Aggregation for Behavior Prediction}, 
  year={2022},
  volume={},
  number={},
  pages={7814-7821},
}

@inproceedings{zeng2021lanercnn,
  title={Lanercnn: Distributed representations for graph-centric motion forecasting},
  author={Zeng, Wenyuan and Liang, Ming and Liao, Renjie and Urtasun, Raquel},
  booktitle={2021 IEEE/RSJ International Conference on Intelligent Robots and Systems (IROS)},
  pages={532--539},
  year={2021},
  organization={IEEE}
}

@inproceedings{zhao2021tnt,
  title={Tnt: Target-driven trajectory prediction},
  author={Zhao, Hang and Gao, Jiyang and Lan, Tian and Sun, Chen and Sapp, Ben and Varadarajan, Balakrishnan and Shen, Yue and Shen, Yi and Chai, Yuning and Schmid, Cordelia and others},
  booktitle={Conference on Robot Learning},
  year={2021},
  organization={PMLR}
}

@inproceedings{gu2021densetnt,
  title={Densetnt: End-to-end trajectory prediction from dense goal sets},
  author={Gu, Junru and Sun, Chen and Zhao, Hang},
  booktitle={Proceedings of the IEEE/CVF International Conference on Computer Vision},
  pages={15303--15312},
  year={2021}
}

@inproceedings{gilles2022gohome,
  title={Gohome: Graph-oriented heatmap output for future motion estimation},
  author={Gilles, Thomas and Sabatini, Stefano and Tsishkou, Dzmitry and Stanciulescu, Bogdan and Moutarde, Fabien},
  booktitle={2022 international conference on robotics and automation (ICRA)},
  pages={9107--9114},
  year={2022},
  organization={IEEE}
}

@inproceedings{liu2021multimodal,
  title={Multimodal motion prediction with stacked transformers},
  author={Liu, Yicheng and Zhang, Jinghuai and Fang, Liangji and Jiang, Qinhong and Zhou, Bolei},
  booktitle={Proceedings of the IEEE/CVF conference on computer vision and pattern recognition},
  pages={7577--7586},
  year={2021}
}

@article{girgis2021latent,
  title={Latent variable sequential set transformers for joint multi-agent motion prediction},
  author={Girgis, Roger and Golemo, Florian and Codevilla, Felipe and Weiss, Martin and D'Souza, Jim Aldon and Kahou, Samira Ebrahimi and Heide, Felix and Pal, Christopher},
  journal={arXiv preprint arXiv:2104.00563},
  year={2021}
}

@inproceedings{huang2022multi,
  title={Multi-modal motion prediction with transformer-based neural network for autonomous driving},
  author={Huang, Zhiyu and Mo, Xiaoyu and Lv, Chen},
  booktitle={2022 International Conference on Robotics and Automation (ICRA)},
  pages={2605--2611},
  year={2022},
  organization={IEEE}
}

@ARTICLE{huang2023differentiable,
  author={Huang, Zhiyu and Liu, Haochen and Wu, Jingda and Lv, Chen},
  journal={IEEE Transactions on Neural Networks and Learning Systems}, 
  title={Differentiable Integrated Motion Prediction and Planning With Learnable Cost Function for Autonomous Driving}, 
  year={2023},
  volume={},
  number={},
  pages={1-15},
  }

@article{shi2022mtr,
  title={Motion transformer with global intention localization and local movement refinement},
  author={Shi, Shaoshuai and Jiang, Li and Dai, Dengxin and Schiele, Bernt},
  journal={Advances in Neural Information Processing Systems},
  volume={35},
  pages={6531--6543},
  year={2022}
}

@ARTICLE{shi2024mtr++,
  author={Shi, Shaoshuai and Jiang, Li and Dai, Dengxin and Schiele, Bernt},
  journal={IEEE Transactions on Pattern Analysis and Machine Intelligence}, 
  title={MTR++: Multi-Agent Motion Prediction With Symmetric Scene Modeling and Guided Intention Querying}, 
  year={2024},
  volume={46},
  number={5},
  pages={3955-3971},
  }

@inproceedings{NIPS2017_3f5ee243,
 author = {Vaswani, Ashish and Shazeer, Noam and Parmar, Niki and Uszkoreit, Jakob and Jones, Llion and Gomez, Aidan N and Kaiser, \L ukasz and Polosukhin, Illia},
 booktitle = {Advances in Neural Information Processing Systems},
 editor = {I. Guyon and U. Von Luxburg and S. Bengio and H. Wallach and R. Fergus and S. Vishwanathan and R. Garnett},
 pages = {},
 publisher = {Curran Associates, Inc.},
 title = {Attention is All you Need},
 volume = {30},
 year = {2017}
}

@InProceedings{Sadeghian_2019_CVPR,
author = {Sadeghian, Amir and Kosaraju, Vineet and Sadeghian, Ali and Hirose, Noriaki and Rezatofighi, Hamid and Savarese, Silvio},
title = {SoPhie: An Attentive GAN for Predicting Paths Compliant to Social and Physical Constraints},
booktitle = {Proceedings of the IEEE/CVF Conference on Computer Vision and Pattern Recognition},
month = {June},
year = {2019}
}

@InProceedings{Chen_2022_CVPR,
    author    = {Chen, Yuxiao and Ivanovic, Boris and Pavone, Marco},
    title     = {ScePT: Scene-Consistent, Policy-Based Trajectory Predictions for Planning},
    booktitle = {Proceedings of the IEEE/CVF Conference on Computer Vision and Pattern Recognition (CVPR)},
    month     = {June},
    year      = {2022},
    pages     = {17103-17112}
}

@InProceedings{Phan-Minh_2020_CVPR,
author = {Phan-Minh, Tung and Grigore, Elena Corina and Boulton, Freddy A. and Beijbom, Oscar and Wolff, Eric M.},
title = {CoverNet: Multimodal Behavior Prediction Using Trajectory Sets},
booktitle = {Proceedings of the IEEE/CVF Conference on Computer Vision and Pattern Recognition (CVPR)},
month = {June},
year = {2020}
}

@InProceedings{Narayanan_2021_CVPR,
    author    = {Narayanan, Sriram and Moslemi, Ramin and Pittaluga, Francesco and Liu, Buyu and Chandraker, Manmohan},
    title     = {Divide-and-Conquer for Lane-Aware Diverse Trajectory Prediction},
    booktitle = {Proceedings of the IEEE/CVF Conference on Computer Vision and Pattern Recognition (CVPR)},
    month     = {June},
    year      = {2021},
    pages     = {15799-15808}
}

@InProceedings{pmlr-v87-casas18a,
  title = 	 {IntentNet: Learning to Predict Intention from Raw Sensor Data},
  author =       {Casas, Sergio and Luo, Wenjie and Urtasun, Raquel},
  booktitle = 	 {Proceedings of The 2nd Conference on Robot Learning},
  pages = 	 {947--956},
  year = 	 {2018},
  editor = 	 {Billard, Aude and Dragan, Anca and Peters, Jan and Morimoto, Jun},
  volume = 	 {87},
  series = 	 {Proceedings of Machine Learning Research},
  month = 	 {29--31 Oct},
  publisher =    {PMLR}
}

@article{rusu2016progressive,
  title={Progressive neural networks},
  author={Rusu, Andrei A and Rabinowitz, Neil C and Desjardins, Guillaume and Soyer, Hubert and Kirkpatrick, James and Kavukcuoglu, Koray and Pascanu, Razvan and Hadsell, Raia},
  journal={arXiv preprint arXiv:1606.04671},
  year={2016}
}

@article{kirkpatrick2017overcoming,
  title={Overcoming catastrophic forgetting in neural networks},
  author={Kirkpatrick, James and Pascanu, Razvan and Rabinowitz, Neil and Veness, Joel and Desjardins, Guillaume and Rusu, Andrei A and Milan, Kieran and Quan, John and Ramalho, Tiago and Grabska-Barwinska, Agnieszka and others},
  journal={Proceedings of the national academy of sciences},
  volume={114},
  number={13},
  pages={3521--3526},
  year={2017},
  publisher={National Acad Sciences}
}

@article{li2017learning,
  title={Learning without forgetting},
  author={Li, Zhizhong and Hoiem, Derek},
  journal={IEEE transactions on pattern analysis and machine intelligence},
  volume={40},
  number={12},
  pages={2935--2947},
  year={2017},
  publisher={IEEE}
}

@article{shumailov2024ai,
  title={AI models collapse when trained on recursively generated data},
  author={Shumailov, Ilia and Shumaylov, Zakhar and Zhao, Yiren and Papernot, Nicolas and Anderson, Ross and Gal, Yarin},
  journal={Nature},
  volume={631},
  number={8022},
  pages={755--759},
  year={2024},
  publisher={Nature Publishing Group UK London}
}

@article{parisi2019continual,
  title={Continual lifelong learning with neural networks: A review},
  author={Parisi, German I and Kemker, Ronald and Part, Jose L and Kanan, Christopher and Wermter, Stefan},
  journal={Neural networks},
  volume={113},
  pages={54--71},
  year={2019},
  publisher={Elsevier}
}

@article{bao2023lifelong,
  title={Lifelong vehicle trajectory prediction framework based on generative replay},
  author={Bao, Peng and Chen, Zonghai and Wang, Jikai and Dai, Deyun and Zhao, Hao},
  journal={IEEE Transactions on Intelligent Transportation Systems},
  year={2023},
  publisher={IEEE}
}

@inproceedings{feng2023continual,
  title={Continual Trajectory Prediction with Uncertainty-Aware Generative Memory Replay},
  author={Feng, Xiushi and Liu, Shuncheng and Chen, Haitian and Zheng, Kai},
  booktitle={2023 IEEE International Conference on Data Mining (ICDM)},
  year={2023},
  organization={IEEE}
}

@article{knoedler2022improving,
  title={Improving pedestrian prediction models with self-supervised continual learning},
  author={Knoedler, Luzia and Salmi, Chadi and Zhu, Hai and Brito, Bruno and Alonso-Mora, Javier},
  journal={IEEE Robotics and Automation Letters},
  volume={7},
  number={2},
  pages={4781--4788},
  year={2022},
  publisher={IEEE}
}

@article{wu2022continual,
  title={Continual pedestrian trajectory learning with social generative replay},
  author={Wu, Ya and Bighashdel, Ariyan and Chen, Guang and Dubbelman, Gijs and Jancura, Pavol},
  journal={IEEE Robotics and Automation Letters},
  volume={8},
  number={2},
  pages={848--855},
  year={2022},
  publisher={IEEE}
}

@article{yang2022continual,
  title={Continual learning-based trajectory prediction with memory augmented networks},
  author={Yang, Biao and Fan, Fucheng and Ni, Rongrong and Li, Jie and Kiong, Loochu and Liu, Xiaofeng},
  journal={Knowledge-Based Systems},
  volume={258},
  pages={110022},
  year={2022},
  publisher={Elsevier}
}

@article{ma2021continual,
  title={Continual multi-agent interaction behavior prediction with conditional generative memory},
  author={Ma, Hengbo and Sun, Yaofeng and Li, Jiachen and Tomizuka, Masayoshi and Choi, Chiho},
  journal={IEEE Robotics and Automation Letters},
  pages={8410--8417},
  year={2021},
  publisher={IEEE}
}

@inproceedings{
ha2017hypernetworks,
title={HyperNetworks},
author={David Ha and Andrew M. Dai and Quoc V. Le},
booktitle={International Conference on Learning Representations},
year={2017},
}

@inproceedings{
Oswald2020Continual,
title={Continual learning with hypernetworks},
author={Johannes von Oswald and Christian Henning and Benjamin F. Grewe and João Sacramento},
booktitle={International Conference on Learning Representations},
year={2020},
}

@inproceedings{rezende2015variational,
  title={Variational inference with normalizing flows},
  author={Rezende, Danilo and Mohamed, Shakir},
  booktitle={International conference on machine learning},
  pages={1530--1538},
  year={2015},
  organization={PMLR}
}

@article{charpentier2020posterior,
  title={Posterior network: Uncertainty estimation without ood samples via density-based pseudo-counts},
  author={Charpentier, Bertrand and Z{\"u}gner, Daniel and G{\"u}nnemann, Stephan},
  journal={Advances in neural information processing systems},
  volume={33},
  pages={1356--1367},
  year={2020}
}

@inproceedings{
charpentier2022natural,
title={Natural Posterior Network: Deep Bayesian Predictive Uncertainty for Exponential Family Distributions},
author={Bertrand Charpentier and Oliver Borchert and Daniel Z{\"u}gner and Simon Geisler and Stephan G{\"u}nnemann},
booktitle={International Conference on Learning Representations},
year={2022},
}

@inproceedings{
ortiz2024magnitude,
title={Magnitude Invariant Parametrizations Improve Hypernetwork Learning},
author={Jose Javier Gonzalez Ortiz and John Guttag and Adrian V Dalca},
booktitle={The Twelfth International Conference on Learning Representations},
year={2024},
}

@inproceedings{
Chang2020Principled,
title={Principled Weight Initialization for Hypernetworks},
author={Oscar Chang and Lampros Flokas and Hod Lipson},
booktitle={International Conference on Learning Representations},
year={2020},
}

@article{dinh2014nice,
  title={Nice: Non-linear independent components estimation},
  author={Dinh, Laurent and Krueger, David and Bengio, Yoshua},
  journal={arXiv preprint arXiv:1410.8516},
  year={2014}
}

@article{kingma2018glow,
  title={Glow: Generative flow with invertible 1x1 convolutions},
  author={Kingma, Durk P and Dhariwal, Prafulla},
  journal={Advances in neural information processing systems},
  volume={31},
  year={2018}
}

@inproceedings{wiederer2023joint,
  title={Joint Out-of-Distribution Detection and Uncertainty Estimation for Trajectory Prediction},
  author={Wiederer, Julian and Schmidt, Julian and Kressel, Ulrich and Dietmayer, Klaus and Belagiannis, Vasileios},
  booktitle={2023 IEEE/RSJ International Conference on Intelligent Robots and Systems (IROS)},
  pages={5487--5494},
  year={2023},
  organization={IEEE}
}

@article{kirichenko2020normalizing,
  title={Why normalizing flows fail to detect out-of-distribution data},
  author={Kirichenko, Polina and Izmailov, Pavel and Wilson, Andrew G},
  journal={Advances in neural information processing systems},
  volume={33},
  pages={20578--20589},
  year={2020}
}

@article{feng2021intelligent,
  title={Intelligent driving intelligence test for autonomous vehicles with naturalistic and adversarial environment},
  author={Feng, Shuo and Yan, Xintao and Sun, Haowei and Feng, Yiheng and Liu, Henry X},
  journal={Nature communications},
  volume={12},
  number={1},
  pages={748},
  year={2021},
  publisher={Nature Publishing Group UK London}
}

@inproceedings{rounDdataset,
    title={The rounD Dataset: A Drone Dataset of Road User Trajectories at Roundabouts in Germany},
    author={Krajewski, Robert and Moers, Tobias and Bock, Julian and Vater, Lennart and Eckstein, Lutz},
    booktitle={2020 IEEE 23rd International Conference on Intelligent Transportation Systems (ITSC)},
    pages={1-6},
    year={2020},
}

@inproceedings{highDdataset,
               title={The highD Dataset: A Drone Dataset of Naturalistic Vehicle Trajectories on German Highways for Validation of Highly Automated Driving Systems},
               author={Krajewski, Robert and Bock, Julian and Kloeker, Laurent and Eckstein, Lutz},
               booktitle={2018 21st International Conference on Intelligent Transportation Systems (ITSC)},
               pages={2118-2125},
               year={2018},
}

@misc{chen2023womdlidar,
      title={WOMD-LiDAR: Raw Sensor Dataset Benchmark for Motion Forecasting}, 
      author={Kan Chen and Runzhou Ge and Hang Qiu and Rami Ai-Rfou and Charles R. Qi and Xuanyu Zhou and Zoey Yang and Scott Ettinger and Pei Sun and Zhaoqi Leng and Mustafa Mustafa and Ivan Bogun and Weiyue Wang and Mingxing Tan and Dragomir Anguelov},
      year={2023},
      eprint={2304.03834},
      archivePrefix={arXiv},
      primaryClass={cs.CV}
}

@article{lester2021power,
  title={The power of scale for parameter-efficient prompt tuning},
  author={Lester, Brian and Al-Rfou, Rami and Constant, Noah},
  journal={arXiv preprint arXiv:2104.08691},
  year={2021}
}

@article{li2021prefix,
  title={Prefix-tuning: Optimizing continuous prompts for generation},
  author={Li, Xiang Lisa and Liang, Percy},
  journal={arXiv preprint arXiv:2101.00190},
  year={2021}
}

@inproceedings{houlsby2019parameter,
  title={Parameter-efficient transfer learning for NLP},
  author={Houlsby, Neil and Giurgiu, Andrei and Jastrzebski, Stanislaw and Morrone, Bruna and De Laroussilhe, Quentin and Gesmundo, Andrea and Attariyan, Mona and Gelly, Sylvain},
  booktitle={International conference on machine learning},
  pages={2790--2799},
  year={2019},
  organization={PMLR}
}

@article{rebuffi2017learning,
  title={Learning multiple visual domains with residual adapters},
  author={Rebuffi, Sylvestre-Alvise and Bilen, Hakan and Vedaldi, Andrea},
  journal={Advances in neural information processing systems},
  volume={30},
  year={2017}
}

@article{hu2022lora,
  title={Lora: Low-rank adaptation of large language models.},
  author={Hu, Edward J and Shen, Yelong and Wallis, Phillip and Allen-Zhu, Zeyuan and Li, Yuanzhi and Wang, Shean and Wang, Lu and Chen, Weizhu and others},
  journal={ICLR},
  volume={1},
  number={2},
  pages={3},
  year={2022}
}

@inproceedings{wang2022learning,
  title={Learning to prompt for continual learning},
  author={Wang, Zifeng and Zhang, Zizhao and Lee, Chen-Yu and Zhang, Han and Sun, Ruoxi and Ren, Xiaoqi and Su, Guolong and Perot, Vincent and Dy, Jennifer and Pfister, Tomas},
  booktitle={Proceedings of the IEEE/CVF conference on computer vision and pattern recognition},
  pages={139--149},
  year={2022}
}

@inproceedings{wang2022dualprompt,
  title={Dualprompt: Complementary prompting for rehearsal-free continual learning},
  author={Wang, Zifeng and Zhang, Zizhao and Ebrahimi, Sayna and Sun, Ruoxi and Zhang, Han and Lee, Chen-Yu and Ren, Xiaoqi and Su, Guolong and Perot, Vincent and Dy, Jennifer and others},
  booktitle={European conference on computer vision},
  pages={631--648},
  year={2022},
  organization={Springer}
}

@inproceedings{smith2023coda,
  title={Coda-prompt: Continual decomposed attention-based prompting for rehearsal-free continual learning},
  author={Smith, James Seale and Karlinsky, Leonid and Gutta, Vyshnavi and Cascante-Bonilla, Paola and Kim, Donghyun and Arbelle, Assaf and Panda, Rameswar and Feris, Rogerio and Kira, Zsolt},
  booktitle={Proceedings of the IEEE/CVF conference on computer vision and pattern recognition},
  pages={11909--11919},
  year={2023}
}

@inproceedings{zhang2020side,
  title={Side-tuning: a baseline for network adaptation via additive side networks},
  author={Zhang, Jeffrey O and Sax, Alexander and Zamir, Amir and Guibas, Leonidas and Malik, Jitendra},
  booktitle={European conference on computer vision},
  pages={698--714},
  year={2020},
  organization={Springer}
}

@inproceedings{gao2023unified,
  title={A unified continual learning framework with general parameter-efficient tuning},
  author={Gao, Qiankun and Zhao, Chen and Sun, Yifan and Xi, Teng and Zhang, Gang and Ghanem, Bernard and Zhang, Jian},
  booktitle={Proceedings of the IEEE/CVF International Conference on Computer Vision},
  pages={11483--11493},
  year={2023}
}

@inproceedings{aljundi2018memory,
  title={Memory aware synapses: Learning what (not) to forget},
  author={Aljundi, Rahaf and Babiloni, Francesca and Elhoseiny, Mohamed and Rohrbach, Marcus and Tuytelaars, Tinne},
  booktitle={Proceedings of the European conference on computer vision (ECCV)},
  year={2018}
}

@article{buzzega2020dark,
  title={Dark experience for general continual learning: a strong, simple baseline},
  author={Buzzega, Pietro and Boschini, Matteo and Porrello, Angelo and Abati, Davide and Calderara, Simone},
  journal={Advances in neural information processing systems},
  volume={33},
  pages={15920--15930},
  year={2020}
}

@inproceedings{
dosovitskiy2021an,
title={An Image is Worth 16x16 Words: Transformers for Image Recognition at Scale},
author={Alexey Dosovitskiy and Lucas Beyer and Alexander Kolesnikov and Dirk Weissenborn and Xiaohua Zhai and Thomas Unterthiner and Mostafa Dehghani and Matthias Minderer and Georg Heigold and Sylvain Gelly and Jakob Uszkoreit and Neil Houlsby},
booktitle={International Conference on Learning Representations},
year={2021},
}

@article{ILSVRC15,
Author = {Olga Russakovsky and Jia Deng and Hao Su and Jonathan Krause and Sanjeev Satheesh and Sean Ma and Zhiheng Huang and Andrej Karpathy and Aditya Khosla and Michael Bernstein and Alexander C. Berg and Li Fei-Fei},
Title = {{ImageNet Large Scale Visual Recognition Challenge}},
Year = {2015},
journal   = {International Journal of Computer Vision (IJCV)},
volume={115},
number={3},
pages={211-252}
}

@inproceedings{ebrahimi2020adversarial,
  title={Adversarial continual learning},
  author={Ebrahimi, Sayna and Meier, Franziska and Calandra, Roberto and Darrell, Trevor and Rohrbach, Marcus},
  booktitle={European conference on computer vision},
  pages={386--402},
  year={2020},
  organization={Springer}
}

@misc{cifar10,
  title={Learning multiple layers of features from tiny images.(2009)},
  author={Krizhevsky, Alex and Hinton, Geoffrey and others},
  year={2009}
}

@article{mnist,
  title={The MNIST database of handwritten digits},
  author={LeCun, Yann},
  journal={http://yann. lecun. com/exdb/mnist/},
  year={1998}
}

@inproceedings{svhn,
  title={Reading digits in natural images with unsupervised feature learning},
  author={Netzer, Yuval and Wang, Tao and Coates, Adam and Bissacco, Alessandro and Wu, Baolin and Ng, Andrew Y and others},
  booktitle={NIPS workshop on deep learning and unsupervised feature learning},
  volume={2011},
  number={5},
  pages={7},
  year={2011},
  organization={Granada}
}

@article{fashion,
  title={Fashion-mnist: a novel image dataset for benchmarking machine learning algorithms},
  author={Xiao, Han and Rasul, Kashif and Vollgraf, Roland},
  journal={arXiv preprint arXiv:1708.07747},
  year={2017}
}

@article{notmnist,
  title={notmnist dataset},
  author={Bulatov, Y.},
  journal={http://yaroslavvb.blogspot.com/2011/
09/notmnist-dataset.html 22},
  year={2011}
}

\begin{IEEEbiography}[{\includegraphics[width=1in,height=1.25in,clip,keepaspectratio]{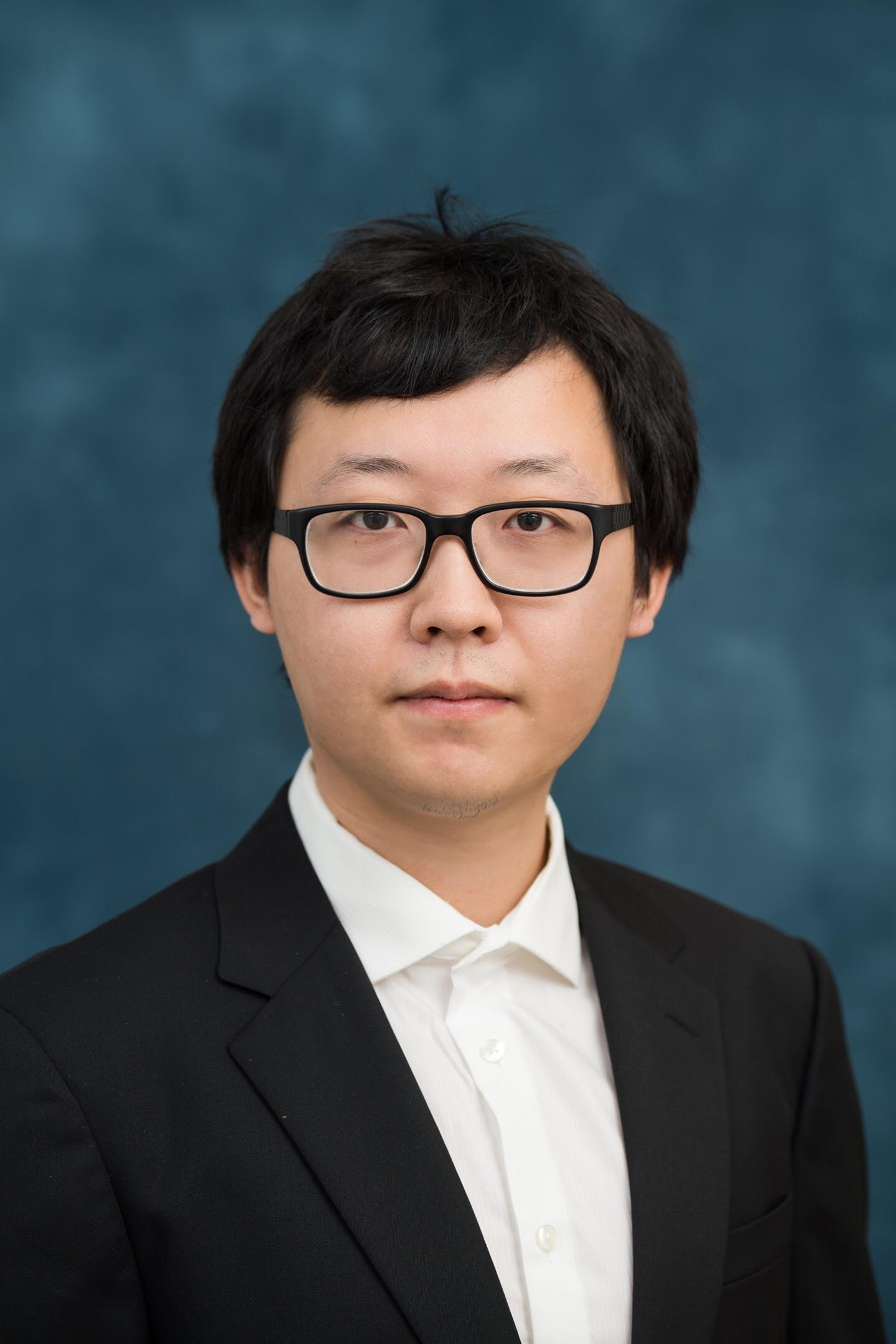}}]{Boqi Li}
received his Ph.D. in Mechanical Engineering from the University of Michigan, Ann Arbor, in 2022. He also holds a B.S. in Mechanical Engineering from the University of Illinois Urbana–Champaign and a M.S. in Mechanical Engineering from Stanford University. His research focuses on the modeling and decision-making for multi-agent systems, particularly for the application of connected, cooperative, and automated mobility systems where the navigation and motion planning of connected and automated vehicles are involved. He is interested in motion prediction, continual learning, and multi-agent reinforcement learning.
\end{IEEEbiography}
 
\vspace{11pt}

\begin{IEEEbiography}[{\includegraphics[width=1in,height=1.25in,clip,keepaspectratio]{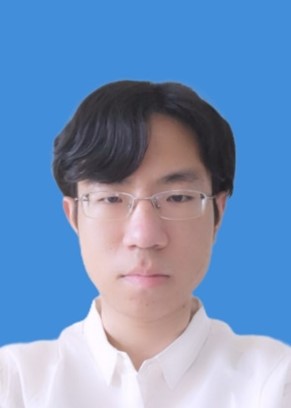}}]{Haojie Zhu}
received the B.S.E degree in electrical and computer engineering from Shanghai Jiao Tong University, Shanghai, China, and the B.S.E degree in mechanical engineering from the University of Michigan, Ann Arbor, MI, USA, in 2020.
He is currently working toward the Ph.D. degree in civil and environmental engineering. His research interests include modeling and motion planning with respect to autonomous vehicles.
\end{IEEEbiography}

\vspace{11pt}

\begin{IEEEbiography}[{\includegraphics[width=1in,height=1.25in,clip,keepaspectratio]{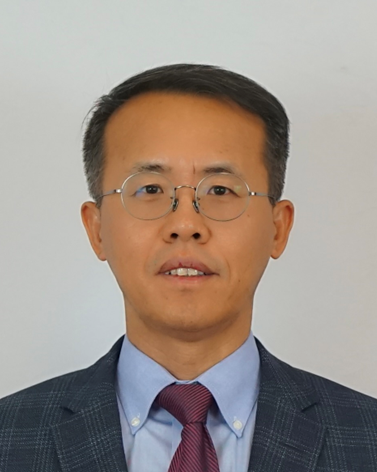}}]{Henry X. Liu} is the Bruce D. Greenshields Collegiate Professor of Engineering and the Director of Mcity at the University of Michigan, Ann Arbor. He is a Professor of Civil and Environmental Engineering, a Professor of Mechanical Engineering, and a Research Professor at the University of Michigan Transportation Research Institute. He also directs the Center for Connected and Automated Transportation, a USDOT funded regional university transportation center. Dr. Liu conducts interdisciplinary research at the interface of transportation engineering, automotive engineering, and artificial intelligence. He is recognized for his foundational work in cyber-physical transportation systems, particularly on the development of smart traffic signal systems with connected vehicles, and testing/evaluation of autonomous vehicles. He has published more than 140 refereed journal articles. His work on safety validation of autonomous vehicles has been published in Nature and featured as the cover story. He has also appeared on a number of media outlets including Wall Street Journal, Forbes, Science Daily, Tech Xplore, CNBC, WXYZ, etc. for transportation innovations. Prof. Liu is the managing editor of Journal of Intelligent Transportation Systems and a board member for the ITS America and IEEE ITS Society.  

\end{IEEEbiography}
\vfill

\end{document}